%% file: main.tex
\documentclass[runningheads]{llncs}
\usepackage{graphicx}

\usepackage{tikz}
\usepackage{comment}
\usepackage{amsmath,amssymb} 
\usepackage{color}
\usepackage{multirow}
\usepackage[accsupp]{axessibility}  

\usepackage{color, colortbl}
\definecolor{LightGray}{rgb}{0.92,0.92,0.92}
\definecolor{Gray1}{rgb}{0.95,0.95,0.95}
\definecolor{Gray2}{rgb}{0.9,0.9,0.9}
\usepackage[ruled,vlined]{algorithm2e}

\SetCommentSty{mycommfont}
\usepackage[title]{appendix}
\usepackage[pagebackref=true,breaklinks=true,letterpaper=true,colorlinks,bookmarks=false]{hyperref}
\usepackage{makecell}
\usepackage{xcolor} 
\usepackage{pifont}
\usepackage{soul}

\definecolor{darkgreen}{rgb}{0.29, 0.33, 0.13}

\usepackage{pifont}
\newcommand{\xmark}{\color{gray}\ding{55}}%
\newcommand{\objtoken}{${\tiny<}{obj}{\tiny>}$}
\newcommand{\modelname}{UniTAB}

\makeatletter
\DeclareRobustCommand\onedot{\futurelet\@let@token\@onedot}
\def\@onedot{\ifx\@let@token.\else.\null\fi\xspace}

\def\eg{\emph{e.g}\onedot} 
\def\ie{\emph{i.e}\onedot} 
\def\cf{\emph{cf}\onedot} 
\def\etc{\emph{etc}\onedot} \def\vs{\emph{vs}\onedot}

\def\etal{\emph{et al}\onedot}
\makeatother

\usepackage[T1]{fontenc}

\begin{document}
\pagestyle{headings}
\mainmatter
\def\ECCVSubNumber{5226}  

\title{\modelname: Unifying Text and Box Outputs for Grounded Vision-Language Modeling}

\titlerunning{\modelname: Unifying Text and Box Outputs for Grounded VL Modeling}
%
\author{Zhengyuan Yang, Zhe Gan, Jianfeng Wang, Xiaowei Hu, \\Faisal Ahmed, Zicheng Liu, Yumao Lu, Lijuan Wang}
%
\authorrunning{Z. Yang et al.}
%
\institute{Microsoft Cloud and AI\\
\email{\{zhengyang,zhe.gan,jianfw,xiaowei.hu,\\fiahmed,zliu,yumaolu,lijuanw\}@microsoft.com}}

\maketitle

\input{abs}
\input{intro}
\input{related}
\input{approach}
\input{exp}
\input{conclusion}

\clearpage

%
%
\bibliographystyle{splncs04}
\bibliography{egbib}

\newpage
\input{supply}

\end{document}

%% file: abs.tex
\begin{abstract}
We propose \modelname~that Unifies Text And Box outputs for grounded vision-language (VL) modeling. Grounded VL tasks such as grounded captioning require the model to generate a text description and align predicted words with object regions. To achieve this, models must generate desired text and box outputs together, and meanwhile indicate the alignments between words and boxes. In contrast to existing solutions that use multiple separate modules for different outputs, \modelname~represents both text and box outputs with a shared token sequence, and introduces a special \objtoken~token to naturally indicate word-box alignments in the sequence. \modelname~thus could provide a more comprehensive and interpretable image description, by freely grounding generated words to object regions. On grounded captioning, \modelname~presents a simpler solution with a single output head, and significantly outperforms state of the art in both grounding and captioning evaluations. On general VL tasks that have different desired output formats (\ie, text, box, or their combination), \modelname~with a single network achieves better or comparable performance than task-specific state of the art. Experiments cover $7$ VL benchmarks, including grounded captioning, visual grounding, image captioning, and visual question answering. Furthermore, \modelname's unified multi-task network and the task-agnostic output sequence design make the model parameter efficient and generalizable to new tasks.\footnote{Code is available at \url{https://github.com/microsoft/UniTAB}.}
\end{abstract}

%% file: intro.tex
\section{Introduction}
\setcounter{footnote}{-1}
\begin{figure*}[t]
\centering
\includegraphics[width=.95\textwidth]{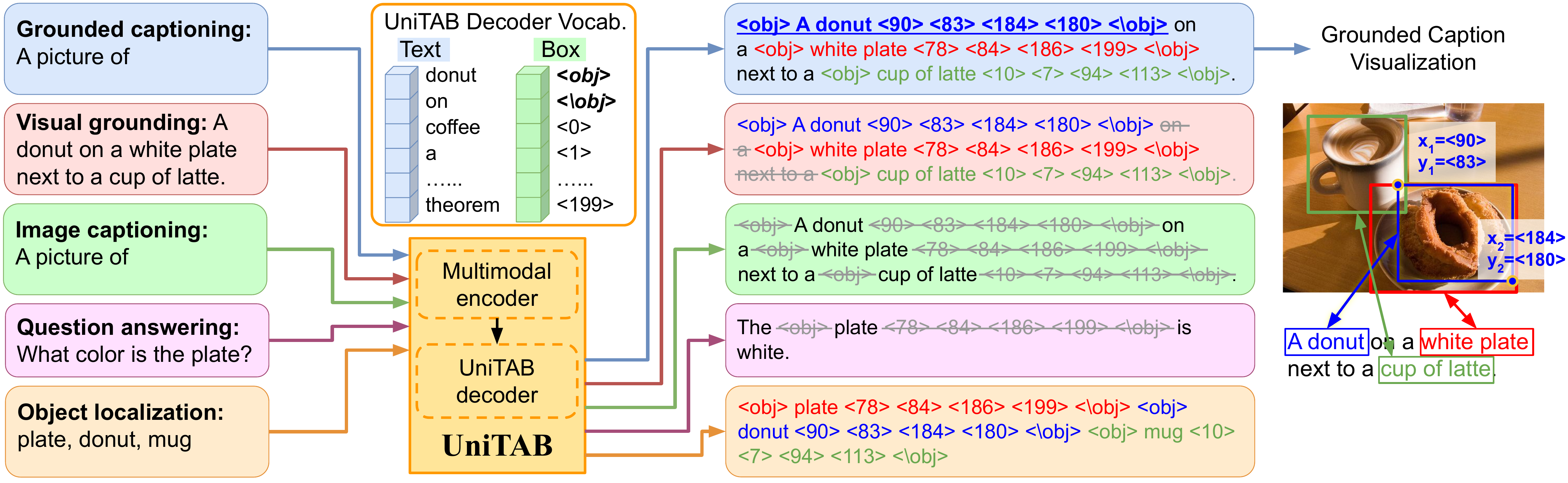}
    \caption[Caption for LOF]{
    We propose \modelname~that Unifies Text And Box outputs with no format-specific modules. \modelname~generates both text and box tokens in an auto-regressive manner, conditioned on the multimodal image-text inputs. The introduced $\tiny{<}obj\tiny{>}$ token naturally indicates the word-box alignments, as shown in word-box pairs of the same color in the right visualization. \modelname~thus can approach a wide range of VL tasks, including the challenging grounded captioning, with a single unified architecture. The \textcolor{gray}{\st{gray}} tokens in the task-agnostic output sequence are predictions not used for downstream task evaluation, \eg, box tokens in image captioning and VQA.
	}
\vspace{-0.14in}
\label{fig:intro}
\end{figure*}
Text sequences~\cite{chen2015microsoft,VQA_15} and bounding boxes~\cite{lin2014microsoft,yu2016modeling} are two representative output formats for image understanding tasks~\cite{deng2009imagenet,lin2014microsoft,chen2015microsoft}. Text is well suited for generating image-level predictions, such as describing an image with a sentence~\cite{chen2015microsoft} or tagging an image with keywords~\cite{fu2017advances}, but fails to refer to a dense image region. On the other hand, box could point to any image area~\cite{lin2014microsoft}, but alone has a limited ability to provide semantically-rich descriptions. A natural question is \emph{can we have a single model that unifies text and box outputs,} \ie, generating both text and box outputs while aligning predicted words with boxes. Unifying these two output formats allows the model to better express its understanding of the image. Taking captioning as an example, such a unified model could ground all noun entities~\cite{zhou2019grounded,plummer2015flickr30k} in the caption back to aligned image regions, thus providing a more comprehensive and interpretable image description. This problem is known as grounded captioning~\cite{zhou2019grounded,zhou2020more,ma2019learning,plummer2015flickr30k}.  Moreover, unifying output formats is one important step toward the grand vision of building task-agnostic, general-purpose vision systems~\cite{Gupta2021TowardsGP} that are parameter efficient and well generalizable.

Recent works~\cite{cho2021unifying,Gupta2021TowardsGP,zhou2019grounded,zhou2020more,ma2019learning} have developed models that can generate both text and box outputs. Specifically, the system combines an online~\cite{Gupta2021TowardsGP} or offline~\cite{cho2021unifying,zhou2019grounded,zhou2020more,ma2019learning} object detection module that predicts boxes, with a vision-language model that generates text. The word and box alignments are then separately generated as additional predictions, such as the relevance score~\cite{Gupta2021TowardsGP,zhou2019grounded,zhou2020more,ma2019learning}. Predicting text, box, and their alignments separately weakens the benefits of a unified system. The separate modules prevent the framework from being simple and parameter efficient. Furthermore, the explicit object detection component increases the model running time~\cite{kim2021vilt} and potentially limits its generalization ability given the preset detector vocabulary~\cite{wang2021simvlm}, as discussed in previous VL studies~\cite{kim2021vilt,wang2021simvlm}. Going beyond these successful initial explorations, we ask a bolder question: \emph{can we unify the output formats with no separate modules}? Specifically, we explore \textbf{1).} how to have a single architecture without an explicit detector jointly generating text and box, and \textbf{2).} how to represent the word-box alignments naturally in the output to avoid the additional alignment prediction. To this end, we model both text and box predictions as an auto-regressive token generation task, and present a single encoder-decoder model that is fully shared among text, box, and alignment predictions.

Our modeling of box prediction takes inspiration from Pix2seq~\cite{chen2021pix2seq}, an object detection study showing that predicting boxes in an auto-regressive manner yields good detection performance~\cite{lin2014microsoft}. Its core idea is to quantize the four coordinates in each box into four discrete box tokens, and arrange them with a fixed order into a token sequence, \ie, $\left[ y_{\text{min}},x_{\text{min}},y_{\text{max}},x_{\text{max}}\right]$. Box prediction can then be modeled as a multi-step classification task, instead of conventional coordinate regression~\cite{girshick2014rich,redmon2016you,carion2020end}. The same classification modeling as in text generation~\cite{radford2018improving} makes it possible to unify text and box prediction. However, Pix2seq is designed for the single-modal object detection task, and does not support open-ended text generation nor multimodal inputs and outputs. Moreover, it is unclear how the text and box alignment is intended to be presented in a unified sequence.

In this study, we propose \modelname~that unifies text and box outputs. As shown in Figure~\ref{fig:intro}, we unify open-ended text generation~\cite{radford2018improving} and discrete box token prediction~\cite{chen2021pix2seq} into a single shared decoder. During the auto-regressive decoding, \modelname~switches to box tokens right after any text words to be grounded, and switches back to text tokens after predicting the box. In \modelname, we study how to handle such text-box code-switching~\cite{enwiki:1068820985} and naturally represent word-box alignments. We introduce a special $\tiny{<}obj\tiny{>}$ token inserted before the text word to be grounded, and after the generated box tokens. The $\tiny{<}obj\tiny{>}$ token simplifies the sequence generation by providing hints of the code-switching, and naturally represents word-box alignments. That is, the words and box within a pair of $\tiny{<}obj\tiny{>}$ tokens refer to the same entity, as shown in word-box pairs of the same color in Figure~\ref{fig:intro}. With the $\tiny{<}obj\tiny{>}$ token and output sequence design, \modelname~approaches grounded VL tasks such as grounded captioning~\cite{zhou2019grounded,plummer2015flickr30k} and phrase grounding~\cite{plummer2015flickr30k} with a single decoder, in contrast to separately predicting text, box, and their alignments with multiple output heads~\cite{zhou2019grounded,ma2019learning,zhou2020more,kamath2021mdetr}.

We further apply \modelname~on general VL tasks~\cite{zhou2019grounded,yu2016modeling,mao2016generation,plummer2015flickr30k,VQA_15,chen2015microsoft,zhou2016learning} and observe two unique properties. \emph{First}, the unified architecture for text, box, and alignment predictions enables \modelname~to perform multi-task training~\cite{aghajanyan2021muppet,wei2021finetuned,aribandi2021ext5}, which learns a single set of parameters for different VL tasks without introducing task-specific heads. Multi-task training avoids task-specific model copies and thus saves the parameters to store. It also facilities the use of data in different tasks, thus boosting the performance of certain VL tasks. \emph{Second}, as shown in Figure~\ref{fig:intro}, \modelname's output sequence is designed to be task-agnostic and shares the same text+box design across different VL tasks. The task-agnostic output design could help \modelname~generalize to certain unseen tasks, by reformatting new tasks' desired outputs into the seen text+box sequences.

We evaluate \modelname~on $7$ VL benchmarks, including grounded captioning~\cite{zhou2019grounded,plummer2015flickr30k}, visual grounding~\cite{yu2016modeling,mao2016generation,plummer2015flickr30k}, image captioning~\cite{chen2015microsoft}, and visual question answering~\cite{VQA_15}, all with a single encoder-decoder network architecture, trained by the cross-entropy language modeling objective~\cite{radford2018improving}. With a unified framework and minimum task-specific assumptions, our model achieves better or comparable performance with task-specific state of the art. In grounded captioning, \modelname~not only presents a simpler solution by eliminating separate task-specific heads~\cite{zhou2019grounded,ma2019learning,zhou2020more,kamath2021mdetr}, but also significantly outperforms the prior art~\cite{ma2019learning,chen2021distributed} (from $62.5$ to $69.7$ in captioning CIDEr score and from $8.44$ to $12.95$ in grounding F1 score). Our contributions are summarized as follows.

\begin{itemize}
\item \modelname~is the first grounded VL model that can approach a wide range of tasks, including the challenging grounded captioning, without separate output modules. We introduce the \objtoken~token that helps text and box outputs synergistically work together, with their alignments naturally represented.
\item \modelname~achieves better or comparable performance to state of the art on $7$ VL benchmarks. Its unified multi-task network and the task-agnostic output sequence design make it parameter efficient and generalizable to new tasks.
\end{itemize}

%% file: related.tex
\section{Related Work}
\input{tabs/related}
\noindent\textbf{Grounded captioning.}
The grounded captioning task~\cite{zhou2019grounded,plummer2015flickr30k} requires the model to generate a text caption and grounds all mentioned noun phrases~\cite{zhou2019grounded,plummer2015flickr30k} to aligned image regions. The input is a single image, and the desired outputs are the caption sentence, multiple object boxes, and the word-box alignments. Existing methods~\cite{zhou2019grounded,ma2019learning,zhou2020more,chen2021distributed} adopt separate output heads for text, box (usually with an offline detector~\cite{ren2015faster,anderson2018bottom}), and alignment predictions. In contrast, \modelname~uses a single decoding sequence to represent all desired outputs.

\noindent\textbf{Vision-language pre-training (VLP).}
Large-scale VLP has become the new training paradigm for VL research. Prior works~\cite{lu2019vilbert,li2019visualbert,alberti2019fusion,li2020unicoder,tan2019lxmert,su2019vl,zhou2020unified,chen2019uniter,lu202012,li2020oscar} first show the power of VLP by using region features obtained from an off-the-shelf object detector~\cite{ren2015faster}. However, the region feature extraction significantly increases the model's computation cost and run time. Recent studies~\cite{huang2020pixel,kim2021vilt,li2021align,wang2021simvlm} shift the paradigm and show that grid features extracted from raw image patches also work well. Most studies adopt similar output architectures of either discriminative classification heads or auto-regressive text decoders. As shown in the second row of Table~\ref{table:related}, these output structures often contain task-specific designs and do not support bounding box prediction, which is an important output format for VL tasks such as visual grounding and grounded image captioning.

\noindent\textbf{Unified VL framework.}
Prior works have presented successful explorations on building VL models with unified input-output formats. VL-T5~\cite{cho2021unifying} and GPV~\cite{Gupta2021TowardsGP} first represent images as object region features with an online or offline object detector~\cite{ren2015faster,carion2020end}. Bounding box prediction is then simplified as index classification over the set of region candidates generated by the detector. The other threads, MDETR~\cite{kamath2021mdetr} and UniT~\cite{hu2021unit}, add task-specific classification heads on top of the DETR object detector~\cite{carion2020end} to perform VL tasks. However, different tasks still require different output heads. Moreover, it is unclear how to extend the framework for open-ended text generation, thus supporting VL tasks like image captioning. In this study, we aim to build a single unified framework that takes structured inputs (\ie, raw image and language) in, and generates structured outputs (\ie, text and boxes), with no output format specific modules.

%% file: tabs/related.tex
\begin{table*}[t]\small
\centering
\caption{Summary of unified VL models. We highlight the desired modeling in \textcolor{blue}{blue}. 
\emph{Visual Modeling:} instead of using an object detection (OD) module, we take raw ``image patches'' as visual input. \emph{Text Output:} instead of using task-specific output heads~\cite{lu2019vilbert,huang2020pixel,kamath2021mdetr,hu2021unit} for different VL tasks (classification or text generation heads), we use a ``single output sequence''~\cite{cho2021unifying,Gupta2021TowardsGP} to approach different tasks. \emph{Box Output:} many prior models cannot predict boxes~\cite{huang2020pixel} or simplify it as region index prediction with detector-generated region proposals~\cite{lu2019vilbert,cho2021unifying,Gupta2021TowardsGP}. We aim to predict ``box coordinates'' without an explicit OD module~\cite{kamath2021mdetr,hu2021unit}. \emph{Word-box Align:} most models fail to generate either open-ended text~\cite{lu2019vilbert,kamath2021mdetr,hu2021unit} or object boxes~\cite{huang2020pixel}, thus cannot represent word-box alignments. In contrast to the extra alignment predictions~\cite{Gupta2021TowardsGP,cho2021unifying}, our introduced $\tiny{<}obj\tiny{>}$ token naturally indicates word-box alignments ``inline'' in the output sequence.
}
\resizebox{1.\textwidth}{!}
{
\begin{tabular}{ l | c c c c}
    \hline
    Representative Models & Visual Modeling & Text Output & Box Output & Word-box Align  \\
    \hline
    {ViLBERT~\cite{lu2019vilbert}, OSCAR~\cite{li2020oscar},} & \multirow{3}{*}{Offline OD} & \multirow{3}{*}{Task-specific Heads} & \multirow{3}{*}{Region Index} & \multirow{3}{*}{\xmark} \\
    {UNITER~\cite{chen2019uniter}, VinVL~\cite{zhang2021vinvl}, } & & & \\
    {\etc~\cite{li2019visualbert,tan2019lxmert,li2020unicoder,su2019vl,zhou2020unified,lu202012}} & & & \\
    \hline
    {PixelBERT~\cite{huang2020pixel}, SOHO~\cite{huang2021seeing},} & \multirow{3}{*}{\textcolor{blue}{Image Patches}} & \multirow{3}{*}{Task-specific Heads} & \multirow{3}{*}{\xmark} & \multirow{3}{*}{\xmark} \\
    {ViLT~\cite{kim2021vilt}, SimVLM~\cite{wang2021simvlm},} & & & \\
    {\etc~\cite{shen2021much,li2021align,xue2021probing,dou2021empirical,wang2022git}} & & & \\
    \hline
    {VL-T5~\cite{cho2021unifying}} & Offline OD & \multirow{2}{*}{\textcolor{blue}{Single Output Seq.}} & \multirow{2}{*}{Region Index} & \multirow{2}{*}{Extra Prediction} \\
    {GPV~\cite{Gupta2021TowardsGP}} & Online OD &  &  &  \\
    \hline
    {MDERT~\cite{kamath2021mdetr}}, {UniT~\cite{hu2021unit}} & \textcolor{blue}{Image Patches} & Task-specific Heads & \textcolor{blue}{Box Coordinate} & \xmark \\
    \hline
    {\modelname~(Ours)} & \textcolor{blue}{Image Patches} & \textcolor{blue}{Single Output Seq.} & \textcolor{blue}{Box Coordinate} & \textcolor{blue}{Inline Indicated}\\
    \hline
\end{tabular}
}
\label{table:related}
\end{table*}

%% file: approach.tex
\begin{figure*}[t]
\centering
\includegraphics[width=.95\textwidth]{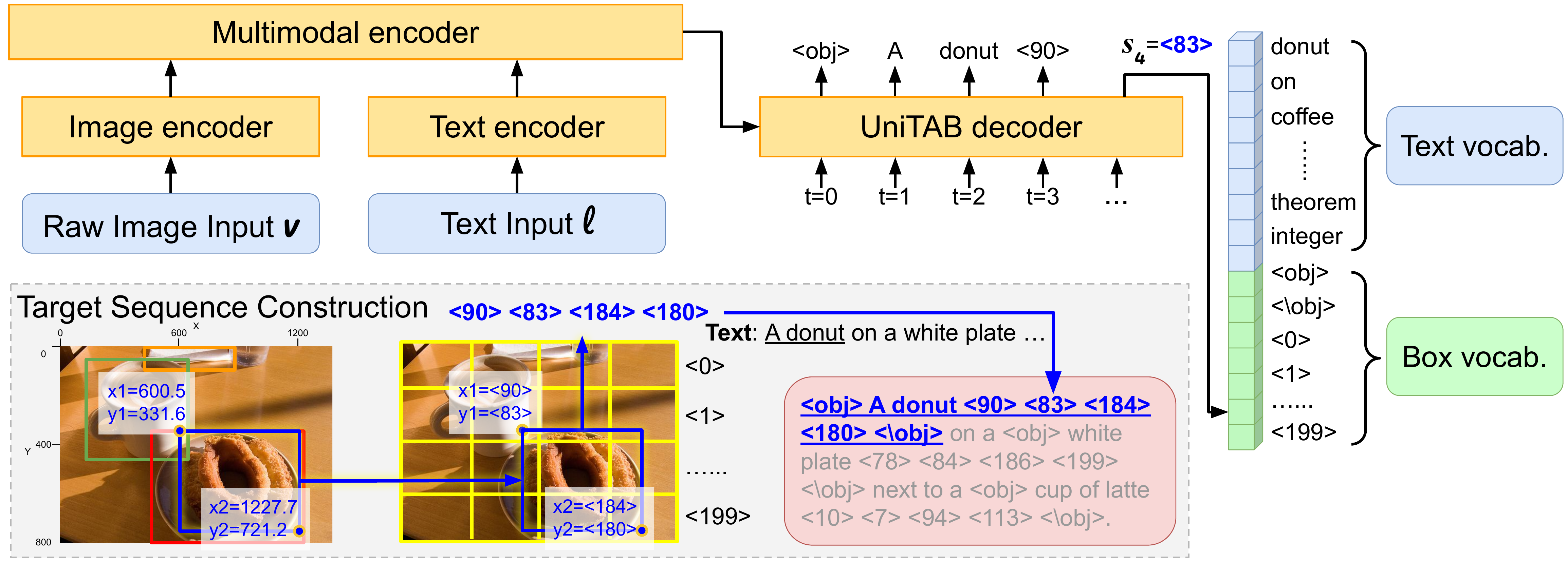}
    \caption{
    \modelname~is an encoder-decoder framework that can jointly output open-ended text and box without output format specific modules. A transformer encoder-decoder takes the encoded image-text features to predict the target text+box sequence. The bottom sub-figure illustrates the output target sequence design. We introduce a special $\tiny{<}obj\tiny{>}$ token to indicate the alignments between predicted words and boxes, such as words ``\textcolor{blue}{a donut}'' and the \textcolor{blue}{blue} box. During decoding, the output sequence could seamlessly switch between text and box tokens to ground an object, if applicable.
	}
\label{fig:arch}
\end{figure*}

\section{The \modelname~Framework}
\subsection{Architecture Overview}
We implement \modelname~using a transformer encoder-decoder architecture built on top of the single-modality image and text encoders, as shown in Figure~\ref{fig:arch}. For image, we use ResNet-101~\cite{he2016deep} to encode the raw image input $v$, and flatten the grid features as the visual representation. For text, we use RoBERTa$_\text{BASE}$~\cite{liu2019roberta} to encode input text $l$ into hidden word features. The encoded image and text features are then projected into a shared embedding space. We use a $6$-layer transformer encoder that takes the concatenated image and text feature sequence as input, and a $6$-layer transformer decoder for output sequence generation. The decoder generates output tokens in an auto-regressive manner, similar to language modeling~\cite{radford2018improving,raffel2019exploring}. The \modelname~decoder could generate tokens from both the text and box vocabularies, as shown in the right part of Figure~\ref{fig:arch}.

\subsection{\modelname~Target Output Sequence}
We show how to construct ground-truth target output sequences, such that text and box can be jointly represented with word-box alignments contained inline.

\noindent\textbf{Box token sequence.}
We first review the bounding box quantization approach introduced in Pix2seq~\cite{chen2021pix2seq}. As shown in the bottom part of Figure~\ref{fig:arch}, a rectangular bounding box in a 2D image can be represented by four floating-point numbers, namely $\left[x_\text{min},y_\text{min}, x_\text{max},y_\text{max} \right]$. The established object detection paradigm~\cite{ren2015faster,redmon2016you,carion2020end} predicts four continuous floating-point values to regress the coordinates in a single step. In contrast, Pix2seq quantizes each coordinate into one of the $n_\text{bins}$ discrete bins, and represent each box with four tokens arranged sequentially. We adopt the similar idea and represent each box as four discrete tokens, $\left[\tiny{<}x_\text{min}\tiny{>},\tiny{<}y_\text{min}\tiny{>}, \tiny{<}x_\text{max}\tiny{>},\tiny{<}y_\text{max}\tiny{>}\right]$, where $\tiny{<}x\tiny{>}$,$\tiny{<}y\tiny{>}$ are quantized box tokens ranging from $\tiny{<}0\tiny{>}$, to $\tiny{<}n_\text{bins}-1\tiny{>}$.

\noindent\textbf{Unified decoding sequence with $\mathbf{\tiny{<}obj\tiny{>}}$ token.}
We aim to have a unified decoding sequence $s$ that can jointly represent text and box, meanwhile indicating word-box alignments. For the former, we unify the text and box vocabularies such that a single decoder can freely generate text or box tokens at any decoding step. Specifically, \modelname's decoding vocabulary contains both text and box tokens, and has a size of $n_\text{text}+n_\text{bins}+2$. $n_\text{text}$ and $n_\text{bins}$ are the text vocabulary size and the number of coordinate bins. We use the same set of $n_\text{bins}$ box tokens~\cite{chen2021pix2seq} for all four box coordinates. The output token selection at each decoding step is conducted over the entire unified vocabulary.

The remaining question is how to represent the word-box alignments in the output sequence. Instead of extra alignment score prediction~\cite{Gupta2021TowardsGP,zhou2019grounded,zhou2020more,ma2019learning}, we represent word-box alignments inline with two introduced special tokens $\tiny{<}obj\tiny{>}$ and $\tiny{<}\backslash obj\tiny{>}$. Specifically, the model switches to box tokens right after any text words to be grounded, and inserts the $\tiny{<}obj\tiny{>}$ tokens before the first text word and after the last box tokens, respectively. For example, in Figure~\ref{fig:arch}, we extend the text phrase ``a donut'' in the text-only caption as ``$\tiny{<}obj\tiny{>}$ a donut $\tiny{<}90\tiny{>}$ $\tiny{<}83\tiny{>}$ $\tiny{<}184\tiny{>}$ $\tiny{<}180\tiny{>}$ $\tiny{<}\backslash obj\tiny{>}$'' in the extended target sequence, where $90,83,184,180$ are the quantized box coordinates for the blue box. The word-box alignments then can be easily extracted from the predicted sequence, \ie, words and box within the pair of $\tiny{<}obj\tiny{>}$ tokens refer to the same entity, such as ``a donut.''

\subsection{\modelname~Training}
\label{sec:train}
\noindent\textbf{Objective.}
We train the model with a single language modeling objective~\cite{radford2018improving}, \ie, at each decoding step $t$, maximizing the likelihood of target token $s_t$ conditioned on input image $v$, input text $l$, and previous target tokens $s_{<t}$:
\begin{equation}
\label{eq:lm}
    \mathcal{L}_{LM}(\theta) = -\textstyle{\sum_{t=1}^T}\log P_{\theta}(s_t|s_{<t},v,l),
\end{equation}
where $\theta$ denotes the model parameters, and $T$ is the target sequence length.

\noindent\textbf{Training stages.}
\modelname's unified structure facilitates the pre-training and finetuning that use the same language modeling objective. We train \modelname~with up to three stages. The first is vision-language pre-training, which leverages large-scale image-text dataset optionally with grounded box annotations. Then, we perform multi-task finetuning, where multiple downstream task datasets with supervised annotations are merged to finetune a single model for different VL tasks. Lastly, we could conduct task-specific finetuning that adapts the model to each specific task for further improvement. The three stages share the same training objective as in Eq.~\ref{eq:lm}, but with different training corpus and input-output designs. We discuss the combinations of these different training stages in Section~\ref{sec:ablation}. We next introduce each of these three training stages.

\noindent\textbf{1. Pre-training.}
Pre-training aims to use large-scale data loosely related to downstream tasks for general VL representation learning. We pre-train the model with a single language modeling objective to predict the target sequence $s$, conditioned on image $v$ and input text $l$. We randomly set the input text $l$ as an empty string or the text-only image description, with the same probability of $0.5$. We train the model to generate the text+box sequence $s$ shown in Figure~\ref{fig:arch}. The model thus learns to perform both captioning-like (with empty string input) and grounding-like (with image description input) VL tasks during pre-training.

\noindent\textbf{2. Multi-task finetuning.}
Multi-task finetuning~\cite{aghajanyan2021muppet,wei2021finetuned,aribandi2021ext5} aims to use supervised annotations from multiple downstream task datasets to train a single model, thus avoiding task-specific model copies and further boosting the model performance. \modelname's unified architecture and training objective facilitate the unique property of multi-task finetuning. Instead of having multiple duplicates of a pre-trained model, each optimized for a downstream task, multi-task finetuning trains a single set of parameters to perform all different VL tasks. We gather supervised data annotations from all $7$ experimented VL tasks and train a single model with the language modeling objective. One major advantage of multi-task finetuning is that a single model can support multiple VL tasks, thus saving model parameters. Multi-task finetuning could also improve certain downstream tasks' performance by using annotations from different tasks.

\noindent\textbf{3. Task-specific finetuning.}
\modelname~can also perform the standard task-specific finetuning as in VLP studies~\cite{lu2019vilbert,chen2019uniter,li2020oscar}. Furthermore, we observe that multi-task finetuning not only generates a single model that performs well in different VL tasks, but also serves as a good initialization point for a second-stage task-specific finetuning. We refer to this setting as ``pre-finetuning''~\cite{aghajanyan2021muppet,wei2021finetuned,aribandi2021ext5}.

\noindent\textbf{Inference.}
We use $\arg\max$ sampling to obtain the sequence prediction. We then extract the text and box predictions from the sequence offline for final evaluation. For example, we discard box tokens to get the text prediction, and dequantize box tokens to get the box prediction. Finally, we evaluate the model on each downstream task with its desired output formats, \eg, text for VQA, boxes for visual grounding, or both text and boxes for grounded captioning. We show in Section~\ref{sec:ablation} that the task-agnostic output sequence design could help \modelname~generalize to unseen tasks that require text or box outputs.

%% file: exp.tex
\section{Experiments}
\subsection{Experiment Overview}
\noindent\textbf{Downstream tasks.}
We evaluate \modelname~on $7$ VL benchmarks (later summarized in Table~\ref{table:main}). We start with grounded captioning~\cite{zhou2019grounded,plummer2015flickr30k} that requires the model to predict text, box, and their alignment. We then benchmark \modelname~on other representative VL tasks, including visual grounding~\cite{yu2016modeling,mao2016generation,plummer2015flickr30k}, COCO image captioning~\cite{chen2015microsoft}, and VQAv2 visual question answering~\cite{VQA_15}. \modelname~approaches a wide range of VL tasks with a single unified architecture. In contrast, prior works require task-specific model designs, making it difficult to work on VL tasks with different desired output formats (text, box, or their combination).

\noindent\textbf{Model variants.}
In addition to the comparison with state of the art, we systematically study the following \modelname~variants with different training stages:
\begin{itemize}
\item \textbf{{Separate-scratch}} conducts task-specific finetuning without pre-training.
\item \textbf{Shared-scratch} conducts multi-task finetuning without pre-training.
\item \textbf{Separate} is first pre-trained and then optimized separately for each downstream task, \ie, the standard pretrain-then-finetune setting in VLP~\cite{lu2019vilbert,chen2019uniter,li2020oscar}.
\item \textbf{Shared} uses multi-task finetuning after pre-training, and shares a single set of parameters for all experimented VL tasks.
\item \textbf{Pre-finetuning} adopts two-stage finetuning from a pre-trained checkpoint. The first stage is multi-task finetuning, followed by task-specific finetuning.
\end{itemize}
We take \modelname$_\text{Pre-finetuning}$ as the default setting and refer to it as \modelname. We report the main ``Pre-finetuning'' results in Section~\ref{sec:results}, and discuss the full results of \modelname~variants in Table~\ref{table:main} and Section~\ref{sec:ablation}.

\noindent\textbf{Training corpus.}
The pre-training corpus~\cite{kamath2021mdetr} aggregates images from Flickr30k Entities~\cite{plummer2015flickr30k}, COCO~\cite{lin2014microsoft,chen2015microsoft}, and Visual Genome (VG)~\cite{krishna2017visual} datasets. Text and grounded box annotations are from the referring expression datasets~\cite{yu2016modeling,mao2016generation}, VG regions, Flickr30k Entity annotations, and the GQA dataset~\cite{hudson2019gqa}. The corpus contains around 200K images and 1.3M image-text pairs with grounded box annotations. 
Optionally, we further add the image-text data with no box annotations from Conceptual Captioning~\cite{sharma2018conceptual} and SBU~\cite{ordonez2011im2text} to pre-training, with settings and results detailed in Section~\ref{sec:ablation}.
For multi-task finetuning, we collect supervised annotations from all $7$ downstream datasets~\cite{zhou2019grounded,plummer2015flickr30k,yu2016modeling,mao2016generation,chen2015microsoft,VQA_15} to jointly train a single model for different tasks.

\noindent\textbf{Implementation details.}
The transformer contains $6$ encoder layers and $6$ decoder layers, with $8$ attention heads and a hidden dimension of $256$ in each layer~\cite{carion2020end}. We use the scale and crop augmentation in DETR~\cite{carion2020end} such that the shortest side is between $480$ and $800$ pixels while the longest at most is $1333$. We pre-train the model for $40$ epochs, and finetune for $20$ epochs in multi-task and task-specific settings. We use a learning rate of $1e^{-4}$ and $2e^{-5}$ for transformer layers and backbones. We train our model with AdamW~\cite{loshchilov2017decoupled} and adopt exponential moving average~\cite{tarvainen2017mean,kamath2021mdetr} with a decay rate of $0.9998$ and a weight decay of $1e^{-4}$. More details are provided in Appendix A.

\subsection{Comparison with Prior Arts}
\label{sec:results}

\noindent\textbf{Grounded captioning.}
\input{tabs/grounded_captioning}
The grounded captioning task~\cite{zhou2019grounded,plummer2015flickr30k} requires the model to generate a caption and ground all generated noun phrases to image regions. The final predictions consist of three parts, \ie, the text caption, visual regions as boxes, and the grounding alignments between words and boxes. Instead of separately predicting those outputs with multiple output heads~\cite{zhou2019grounded,ma2019learning,zhou2020more}, \modelname~naturally represents all desired outputs with a single unified text+box output sequence. Following the established benchmarks~\cite{zhou2019grounded,ma2019learning,zhou2020more} on the Flickr30k Entities dataset, we evaluate ``captioning'' and ``grounding'' separately with the caption metrics~\cite{papineni2002bleu,denkowski2014meteor,anderson2016spice,vedantam2015cider} and grounding F1 scores~\cite{zhou2019grounded}, respectively. The F1 score $F1_{all}$ evaluates grounding as a multi-label classification problem, where a correct prediction contains both the same object word as ground-truth (GT) caption and a larger than $0.5$ IoU with the GT box. We also report $F1_{loc}$ that only computes the grounding score on correctly predicted object words.

Table~\ref{table:groundedcaption} compares our method to state of the art~\cite{zhou2019grounded,ma2019learning,zhou2020more,chen2021distributed}. We observe a significant improvement in the grounding quality, with the F1$_{all}$ score improving from $8.44$ to $12.95$, and F1$_{loc}$ from $22.78$ to $34.79$. \modelname~also achieves a better captioning quality, with the CIDEr score improving from $62.5$ to $69.7$, compared with prior arts~\cite{chen2021distributed}. By exploiting image-text data without box in pre-training, we further boost the CIDEr score from $69.7$ to $74.2$, as detailed in Section~\ref{sec:ablation}.

\begin{figure*}[t]
\centering
\includegraphics[width=1.\textwidth]{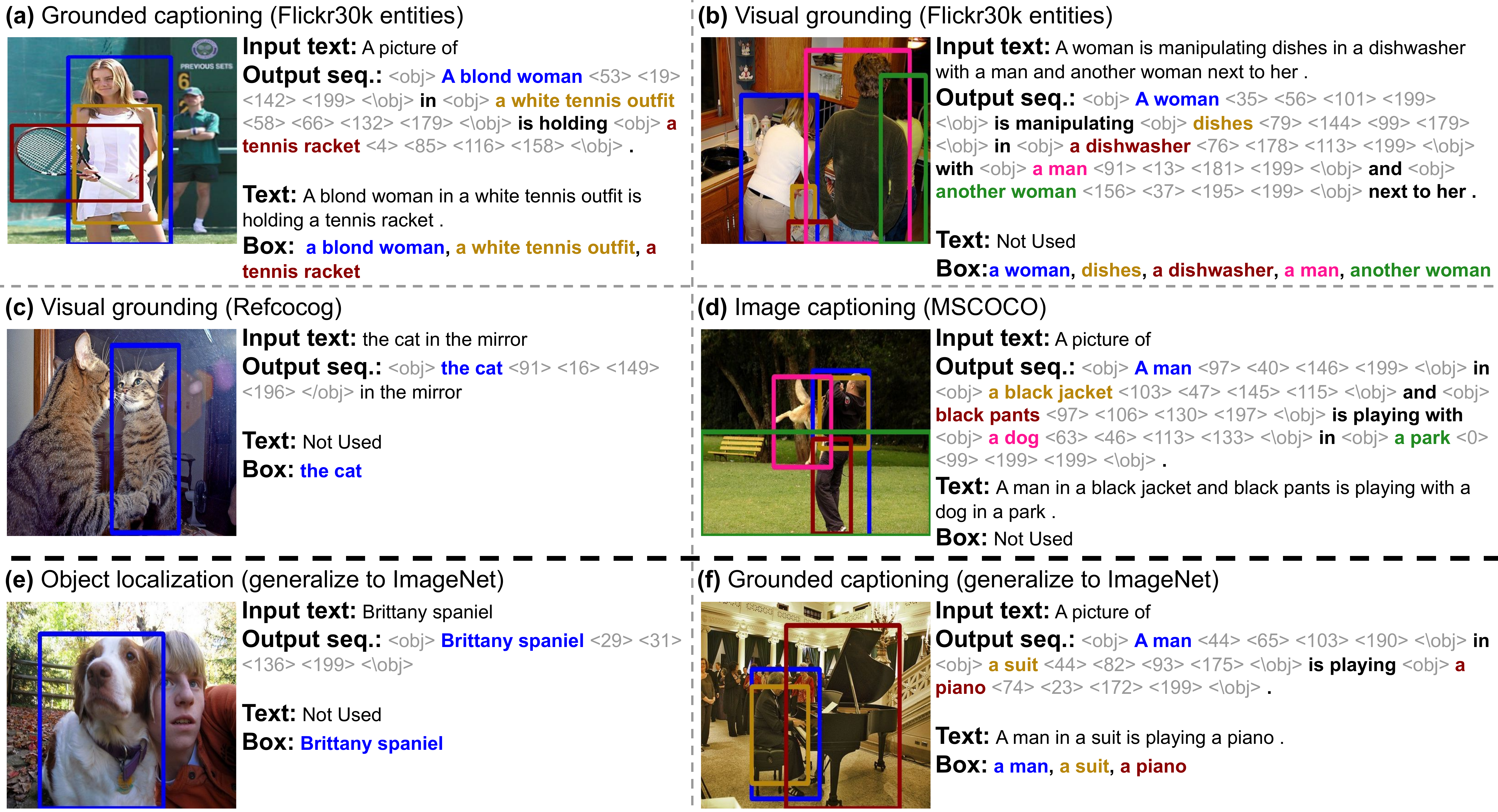}
    \caption{
    Predictions made by \modelname$_\text{Shared}$ that uses a single model for different VL tasks. In each subfigure, we show the input text, the raw output sequence, and the extracted outputs for downstream task evaluations. Specifically, the output sequence contains an open-ended text sequence, box predictions (visualized as bounding boxes), and word-box alignments (visualized as the word-box colors). \textbf{(a-d)} \modelname~approaches a wide range of VL tasks with a single unified model and output sequence. \textbf{(e,f)} With the task-agnostic output sequence, we further generalize \modelname~to unseen images or even new tasks, with examples on ImageNet object localization and grounded captioning.
	}
\label{fig:visu}
\end{figure*}

In addition to the performance improvement, \modelname~presents a simpler and more natural way for the grounded captioning task. Specifically, \modelname~does not require the pre-generated object regions~\cite{zhou2019grounded,ma2019learning,zhou2020more} and avoids using multiple output heads. As shown in Figure~\ref{fig:visu}(a), \modelname~naturally represents text, box, and word-region alignments in a single unified output sequence. Such a simple approach better transfers the model's grounding ability to other datasets or tasks with limited box or grounding annotations, such as COCO caption~\cite{chen2015microsoft} and ImageNet~\cite{deng2009imagenet}, as shown in Figures~\ref{fig:visu}(d,f). We hope \modelname's new paradigm simplifies future studies on grounded VL tasks.

\input{tabs/visual_grounding}

\noindent\textbf{Visual grounding.}
Visual grounding aims to ground language queries into aligned image regions. We experiment on the sub-tasks of referring expression comprehension~\cite{yu2016modeling,mao2016generation} with Refcoco/Refcoco+/Refcocog, and phrase grounding~\cite{plummer2015flickr30k} with Flickr30k Entities. Referring expression comprehension contains a query that describes a single image region and expects a single box prediction. Phrase grounding aims to ground all noun phrases in the input sentence, and requires the model to predict all referred boxes and the word-box alignments. In contrast to previous studies that do not know word-box alignments~\cite{yu2018mattnet,yang2019fast,deng2021transvg} or require separate alignment predictions~\cite{kamath2021mdetr}, \modelname~generates a unified sequence with word-box alignments naturally represented by the special $\tiny{<}obj\tiny{>}$ token. We report the standard metric Acc@0.5~\cite{yu2016modeling,mao2016generation,plummer2015flickr30k}.

As shown in Table~\ref{table:refexp}, \modelname~outperforms the state of the art, including those pre-trained on larger VL corpus~\cite{lu2019vilbert,chen2019uniter,gan2020large} and methods that use carefully-designed task-specific architectures~\cite{yu2018mattnet,yang2019fast,deng2021transvg}.
Moreover, \modelname's unified output with both text and box presents a more natural way of visual grounding, compared to box regression~\cite{yang2019fast,deng2021transvg,kamath2021mdetr} or region index classification~\cite{yu2018mattnet,chen2019uniter,cho2021unifying}.
\modelname's multi-task finetuning enables the use of data from different tasks and datasets, thus boosting performance on all splits, compared with \modelname$_\text{Separate}$.

\noindent\textbf{COCO captioning.}
We benchmark \modelname~on the COCO image captioning dataset~\cite{lin2014microsoft}. We report the results without beam search~\cite{anderson2018bottom} or CIDEr optimization~\cite{rennie2017self}. Table~\ref{table:caption} shows the captioning results on the Karparthy test split~\cite{karpathy2015deep}. We refer to our pre-training corpus as ``200K'' in the ``\#Pre-train'' column, and introduce the corpus used by compared methods later in Appendix A.

\modelname~achieves better performance than prior arts~\cite{xu2021e2e,cho2021unifying} that use similar amounts of pre-training images, with the CIDEr score improved from $117.3$ to $119.8$. Meanwhile, \modelname~does not require input region proposals or object tags~\cite{zhou2020unified,li2020oscar,cho2021unifying}. Using extra image-text pairs~\cite{sharma2018conceptual,ordonez2011im2text} in pre-training further boosts the CIDEr score to $123.1$. We expect a further gain by scaling up the pre-training corpus, as observed in VLP studies~\cite{zhang2021vinvl,li2021align,wang2021simvlm,hu2021scaling}. Despite only being evaluated with caption metrics on COCO, \modelname's unified output sequence could also ground generated noun phrases to image regions, as visualized in Figure~\ref{fig:visu}(d).

\input{tabs/captioning}

\noindent\textbf{Visual question answering.}
\modelname~takes a generative approach to the VQA task~\cite{VQA_15}, where the model generates a free-form text sequence to represent the answer. Table~\ref{table:vqav2} reports the VQA results on both the official test-dev/std split~\cite{VQA_15} and the Karparthy split~\cite{karpathy2015deep} used in VL-T5~\cite{cho2021unifying}. The Karparthy test set is further split into in- and out-domain subsets, based on whether the answer is covered in the top-K (K=$3129$) vocabulary~\cite{cho2021unifying}. The metric is the soft-voting accuracy~\cite{VQA_15}. \modelname~obtains competitive results to the state of the art, and performs better on the Karparthy out-of-domain subset than the discriminative approach~\cite{chen2019uniter}.

\subsection{Ablation and Analysis}
\label{sec:ablation}

\noindent\textbf{Training stage ablation.}
We compare the variants of \modelname~to examine the influence of different pre-training and finetuning stages introduced in Section~\ref{sec:train}. The bottom portion of Table~\ref{table:main} summarizes the results. We first discuss the standard pretrain-then-finetune setting in VLP~\cite{lu2019vilbert,chen2019uniter,li2020oscar} that adopts task-specific finetuning. \textbf{\modelname$_\text{Separate}$} approaches various VL tasks with a single unified architecture, and obtains competitive results to the state of the art that has architectures tailored for each task, or uses larger-scale pre-training data. Compared with \modelname$_\text{Separate-scratch}$ without pre-training, pre-training leads to consistent improvements on all experimented tasks.

\input{tabs/summary_of_results}

With \modelname's unified architecture and output modeling, we can train a single \textbf{\modelname$_\text{Shared}$} model for all experimented VL tasks. Compared with \modelname$_\text{Separate}$, the multi-task finetuning \modelname$_\text{Shared}$ performs comparable or better on experimented VL tasks, while using $\mathbf{7}$ times fewer model parameters by avoiding task-specific model copies. The strong performance of \modelname$_\text{Shared}$ indicates that we can use a single model for multiple downstream tasks, thus being \emph{parameter efficient}. We further experiment with adding task-specific prefixes~\cite{wei2021finetuned,cho2021unifying} to the input text. This variant uses a task-specific prefix such as ``visual grounding:'' to describe each sample's task. We observe that the task prefix has no major influence on model performance, as detailed in Appendix C.

In addition to achieving good performance with a single model, multi-task finetuning \modelname$_\text{Shared}$ also provides a strong initialization point for further task-specific finetuning. \textbf{\modelname$_\text{Pre-finetuning}$} further boosts the performance and achieves better or comparable performance than the state of the art on experimented VL tasks, as shown in the bottom row of Table~\ref{table:main}.

\noindent\textbf{Zero-shot generalization.}
The task-agnostic output sequence design helps \modelname~generalize to new tasks. \modelname~could perform certain tasks in a zero-shot manner by transferring the learned ability of generating text+box sequences $s$ conditioned on image-text inputs. We next explore adapting \modelname~to ImageNet object localization~\cite{deng2009imagenet}. Object localization~\cite{zhou2016learning,choe2020evaluating,wang2021minmaxcam} aims to localize an ImageNet class onto an object region. We take the words in class names as the text input, and have \modelname~generate text+box sequence $s$ conditioned on image-text inputs. We then obtain box predictions by extracting boxes and alignments from $s$, similar to the phrase grounding post-processing. There exist two established benchmark settings. The ``GT-known''~\cite{singh2017hide,zhang2018adversarial,zhang2018self,choe2019attention} setting aims to localize a given ground-truth class. The metrics~\cite{choe2020evaluating} ``MaxBoxAcc'' and ``MaxBoxAccV2'' are the Top-1 accuracy with an IoU threshold of $0.5$, and the average at thresholds $0.3/0.5/0.7$. The second setting tries to localize a predicted class. The metric is ``Top-1 accuracy'' with a $0.5$ IoU threshold. We use EfficientNet~\cite{tan2019efficientnet} classification result with an accuracy of $77.5\%$ for this setting.

We experiment with \modelname$_\text{Shared}$ and show ImageNet object localization results in Table~\ref{table:loc}. \modelname~achieves better performance than the state of the art without using ImageNet images or annotations. The good generalization results show the possibility of generalizing \modelname~to unseen images and tasks in a zero-shot manner. We expect larger-scale pre-training to boost such generalization ability further, as observed in the NLP community~\cite{brown2020language,wei2021finetuned}.
\input{tabs/object_loc}

\noindent\textbf{Pre-training with additional image-text pairs.}
We experiment with adding image-text pairs without boxes in \modelname~pre-training, and examine if the extra image-text data could further improve VL tasks that require text output. For image-text pair data, we pre-train the model to generate the text-only caption conditioned on image and an empty text input. The model variant is referred to as ``Separate$^{\dagger\dagger}$,'' which uses 4M image-text pairs from Conceptual Captioning~\cite{sharma2018conceptual} and SBU~\cite{ordonez2011im2text}. Table~\ref{table:text} compares ``Separate$^{\dagger\dagger}$'' with \modelname$_\text{Separate}$ on grounded captioning, COCO captioning, and VQA. We observe consistent improvements in the text output quality by using extra image-text pairs, \ie, $+8.6$ CIDEr score on grounded captioning~\cite{plummer2015flickr30k}, $+3.8$ CIDEr score on COCO captioning~\cite{chen2015microsoft}, and $+2.5\%$ absolute accuracy on VQA~\cite{VQA_15}. Appendix C further discusses the benefit of pre-training with other addition data, such as boxes from object detection~\cite{lin2014microsoft}.

\noindent\textbf{Model and output sequence design.}
We empirically observe that the introduced $\tiny{<}obj\tiny{>}$ token not only naturally represents the word-box alignment, but also simplifies the sequence prediction by providing hints of the text-box code-switching, thus helping the VL tasks' performance. We postpone the detailed ablation studies on model and output sequence design to Appendix B, including the effectiveness of $\tiny{<}obj\tiny{>}$ token, decoding sampling methods~\cite{anderson2018bottom,holtzman2019curious,chen2021pix2seq}, the number of object tokens, decoding syntactic restrictions, \etc.

\subsection{Qualitative Results}
Figure~\ref{fig:visu} shows the predictions made by \modelname$_\text{Shared}$ on different VL tasks, where all predictions are made by a single model with the same set of parameters. On the right side of each subfigure, we show the input text and predicted output sequence. The output sequence is colored for visualization purposes only, where the text and box colors indicate the word-box alignments. We then show the extracted text and box predictions used for downstream task evaluation. For text, we discard all box tokens to obtain the text-only sequence. For boxes, we keep box tokens and dequantize them as box coordinate predictions~\cite{chen2021pix2seq}.

\modelname's task-agnostic output sequence seamlessly supports different VL tasks. Figure~\ref{fig:visu}(a) shows an example of grounded captioning, where the input text is a blank string and both text and box predictions are used for evaluation. \modelname~could perform the phrase grounding task with the exact output sequence design, by replacing the blank input text with an image description, as shown in Figure~\ref{fig:visu}(b). Figure~\ref{fig:visu}(c) shows a referring expression comprehension example from the Refcocog dataset~\cite{mao2016generation}. The model correctly localizes the referred ``cat'' in the ``mirror.'' Despite not being used by the downstream task evaluation, the model successfully aligns the predicted box with phrase ``the cat.''

\modelname's unified output sequence helps the model transfer the grounded description ability to datasets or tasks with limited box or grounding annotations. As shown in Figure~\ref{fig:visu}(d), \modelname~learns grounded captioning on Flickr30k Entities and transfers such ability to COCO during multi-task finetuning. The generated caption not only has a good caption quality, as evaluated in Table~\ref{table:caption}, but also contains grounding predictions that make the description more comprehensive and interpretable.
With the task-agnostic output sequence design, we further explore generalizing \modelname~to unseen images or even new tasks. Figure~\ref{fig:visu}(e) shows an example of zero-shot object localization on ImageNet. The model correctly localizes the dog conditioned on the text input of ImageNet class label ``brittany spaniel.'' Figure~\ref{fig:visu}(f) shows an example of zero-shot grounded captioning on ImageNet images, where \modelname~generates a smooth caption and correctly grounds all noun phrases. More qualitative results are in Appendix D.

%% file: tabs/grounded_captioning.tex
\begin{table}[t]\small
\centering
\caption{Grounded image captioning results on the test set of Flickr30k Entities~\cite{plummer2015flickr30k}. BLEU@4~\cite{papineni2002bleu}, METEOR~\cite{denkowski2014meteor}, CIDEr~\cite{vedantam2015cider}, and SPICE~\cite{anderson2016spice} metrics are used for caption evaluation. F1$_{all}$ and F1$_{loc}$ metrics~\cite{zhou2019grounded} are used for grounding evaluation. Caption scores with $^{\dagger}$ are optimized with CIDEr~\cite{rennie2017self}.
}
{
\begin{tabular}{ l | c c c c | c c}
    \hline
    \multirow{2}{*}{Method} & \multicolumn{4}{c|}{Caption Eval.} & \multicolumn{2}{c}{~Grounding Eval.~}\\
     & B@4 & M & C & S & F1$_{all}$ & F1$_{loc}$ \\
    \hline
    NBT~\cite{lu2018neural} & 27.1 & 21.7 & 57.5 & 15.6 & - & - \\
    GVD~\cite{zhou2019grounded} & 27.3 & 22.5 & 62.3 & 16.5 & 7.55 & 22.2 \\
    Cyclical~\cite{ma2019learning} & 26.8 & 22.4 & 61.1 & 16.8 & 8.44 & 22.78 \\
    {\footnotesize POS-SCAN~}\cite{zhou2020more} & {30.1}$^{\dagger}$ & 22.6$^{\dagger}$ & {69.3}$^{\dagger}$ & 16.8$^{\dagger}$ & 7.17 & 17.49 \\
    Chen~\etal~\cite{chen2021distributed} & 27.2 & 22.5 & 62.5 & 16.5 & 7.91 & 21.54 \\
    \hline
    \modelname & \textbf{30.1} & \textbf{23.7} & \textbf{69.7} & \textbf{17.4} & \textbf{12.95} & \textbf{34.79} \\
    \hline
\end{tabular}
}
\label{table:groundedcaption}
\end{table}

%% file: tabs/visual_grounding.tex
\begin{table*}[t]\small
\centering
\caption{The performance comparisons (Acc@0.5) on the referring expression comprehension (Refcoco, Refcoco+, Refcocog) and phrase grounding task (Flickr30k Entities). 
}
\begin{tabular}{ l || c c c | c c c | c c || c}
    \hline
    \multirow{2}{*}{Method} &
    \multicolumn{3}{c|}{Refcoco} & \multicolumn{3}{c|}{Refcoco+} & \multicolumn{2}{c||}{Refcocog} & Flickr30k\\
     & val & testA & testB & val & testA & testB & val-u & test-u & Entities \\
    \hline
    MAttNet~\cite{yu2018mattnet} & {76.40} & {80.43} & {69.28} & {64.93} & {70.26} & {56.00} & {66.67} & {67.01} & - \\
    {FAOA~\cite{yang2019fast}} & 72.05 & 74.81 & 67.59 & 55.72 & 60.37 & 48.54 & 59.03 & 58.70 & 68.71 \\
    TransVG~\cite{deng2021transvg} & {81.02} & {82.72} & 78.35 & {64.82} & {70.70} & {56.94} & {68.67} & {67.73} & 79.10 \\
    ViLBERT~\cite{lu2019vilbert} & - & - & - & {72.34} & {78.53} & {62.61} & - & - & - \\
    UNITER~\cite{chen2019uniter} & {81.41} & {87.04} & 74.17 & {75.90} & {81.45} & {66.70} & {74.02} & {68.67} & - \\
    VILLA~\cite{gan2020large} & {82.39} & {87.48} & 74.84 & {76.17} & {81.54} & {66.84} & {76.18} & {76.71} & - \\
    MDETR~\cite{kamath2021mdetr} & {86.75} & {89.58} & 81.41 & {79.52} & {84.09} & {70.62} & {81.64} & {80.89} & \textbf{83.8} \\
    \hline
    \modelname$_\text{Separate}$ & 86.32 & 88.84 & 80.61 & 78.70 & 83.22 & 69.48 & 79.96 & 79.97 & 79.39 \\
    \modelname & \textbf{88.59} & \textbf{91.06} & \textbf{83.75} & \textbf{80.97} & \textbf{85.36} & \textbf{71.55} & \textbf{84.58} & \textbf{84.70} & 79.58 \\
    \hline
\end{tabular}
\label{table:refexp}
\end{table*}

%% file: tabs/captioning.tex
\begin{table}[t]\small
\begin{minipage}{.45\textwidth}
\centering
\caption{COCO image captioning results on the Karparthy test split. The ``\#Pre-train'' column shows the number of pre-training images, if any. }
\resizebox{1.\textwidth}{!}
{
\begin{tabular}{ l | c | c c c c }
    \hline
    Method & \#Pre-train & B@4 & M & C & S \\
    \hline
    Unified VLP~\cite{zhou2020unified} & 3M & 36.5 & 28.4 & 117.7 & 21.3 \\ 
    OSCAR~\cite{li2020oscar} & 4M & {36.5} & {30.3} & {123.7} & {23.1} \\ 
    E2E-VLP~\cite{xu2021e2e} & 180K & {36.2} & - & {117.3} & - \\ 
    VL-T5~\cite{cho2021unifying} & 180K & 34.5 & 28.7 & 116.5 & 21.9 \\ 
    VL-BART~\cite{cho2021unifying} & 180K & 35.1 & 28.7 & 116.6 & 21.5 \\ 
    \hline
    \modelname & 200K & 36.1 & 28.6 & 119.8 & 21.7 \\ 
    \hline
\end{tabular}
}
\label{table:caption}

\end{minipage}%
\qquad
\begin{minipage}{.48\textwidth}

\centering
\caption{Visual question answering results on VQAv2~\cite{VQA_15}. We experiment on both test-dev/test-std splits, and the Karpathy test split used in VL-T5~\cite{cho2021unifying}.}
\resizebox{1.\textwidth}{!}
{
\begin{tabular}{ l | c | c c | c c c }
    \hline
    \multirow{2}{*}{Method} & \#Pre- & \multicolumn{2}{c|}{Test-} & \multicolumn{3}{c}{Karpathy-test} \\
      & train & Dev & Std & In & Out & All \\
    \hline
    UNITER~\cite{chen2019uniter} & 4M & {72.7} & {72.9} & {74.4} & 10.0 & {70.5} \\ 
    VL-T5~\cite{cho2021unifying} & 180K & - & 70.3 & 71.4 & 13.1 & 67.9 \\
    VL-BART~\cite{cho2021unifying} & 180K & - & 71.3 & 72.1 & 13.2 & 68.6 \\
    \hline
    \modelname & 200K & 70.7 & 71.0 & 71.1 & 11.1 & 67.5 \\
    \hline
\end{tabular}
}
\label{table:vqav2}
\end{minipage}
\end{table}

%% file: tabs/summary_of_results.tex
\begin{table*}[t]\small
\fontsize{6}{7}\selectfont
\centering
\caption{{Summary of results obtained by \modelname~and its variants}. The compared methods (upper portion) use task-specific architectures and training objectives, thus could only perform a subset of VL tasks. \modelname~(bottom portion) approaches all tasks with a unified framework and obtains competitive performance. The Refcoco/Refcoco+/Refcocog numbers are on the val set. The Flickr grounding and grounded caption results are on the test set. VQAv2-KP is the VQA Karpathy split~\cite{cho2021unifying}. \modelname$_\text{Pre-finetuning}$ is the default setting that is also referred to as \modelname.
}
\resizebox{1.\textwidth}{!}
{
\begin{tabular}{ l | c || c c c c | c c | c | c c}
    \hline
    \multirow{2}{*}{Method} & \#Pre- &
    \multicolumn{4}{c|}{Visual grounding} & \multicolumn{2}{c|}{Grounded caption} & COCO & \multicolumn{2}{c}{VQAv2}\\
     & train & Refcoco & Refcoco+ & Refcocog & Flickr & Cider & F1$_{all}$ & test-Cider & test-dev & KP-test \\
    \hline
    MDETR~\cite{kamath2021mdetr} & 200K & 86.75 & {79.52} & 81.64 & 83.8 & - & - & - & 70.6 & - \\
    UNITER~\cite{chen2019uniter} & 4M & 81.24 & 75.31 & 74.31 & - & - & - & - & 72.7 & {70.5} \\
    GVD~\cite{zhou2019grounded} & - & - & - & - & - & 62.3 & 7.55 & - & - & - \\
    VL-T5~\cite{cho2021unifying} & 180K & - & - & 71.2 & - & - & - & 116.5 & - & 67.9 \\
    OSCAR~\cite{li2020oscar} & 4M & - & - & - & - & - & - & {123.7} & {73.2} & - \\
    \hline
    \modelname~Variants &  & & & & & & & & & \\
    Separate-scratch & None & 72.96 & 64.98 & 63.56 & 73.40 & 60.5 & 9.22 & 105.3 & 55.4 & 52.4 \\
    Shared-scratch & None & 82.06 & 70.72 & 73.39 & 65.67 & 61.1 & 7.85 & 111.8 & 65.8 & 63.1 \\
    \rowcolor{Gray1}
    Separate & 200K & 86.32 & 78.70 & 79.96 & 79.39 & 65.6 & 11.46 & 119.3 & 69.9 & 66.6 \\
    \rowcolor{Gray1}
    Shared & 200K & 88.50 & 80.98 & 84.46 & 79.23 & 63.4 & 9.18 & 115.8 & 69.1 & 66.6 \\
    \rowcolor{Gray2}
    Pre-finetuning & 200K & 88.59 & 80.97 & 84.58 & 79.58 & 69.7 & 12.95 & 119.8 & 70.7 & 67.5 \\
    \hline
\end{tabular}
}
\label{table:main}
\end{table*}

%% file: tabs/object_loc.tex
\begin{table}[t]\small
\begin{minipage}{.45\textwidth}
\centering
\caption{Zero-shot object localization results on ImageNet~\cite{deng2009imagenet}. Prior works with the weakly supervised setting use ImageNet class labels.}
\resizebox{1.\textwidth}{!}
{
\begin{tabular}{ l c c c}
    \hline
    Method & {\scriptsize Top-1 Acc.} & {\scriptsize MaxBoxAcc} & {\scriptsize MaxBoxAccV2} \\
    \hline
    CAM~\cite{zhou2016learning} & 51.8 & 64.2 & 63.7 \\
    HaS~\cite{singh2017hide} & 49.9 & 63.1 & 63.4 \\
    CutMix~\cite{yun2019cutmix} & 51.5 & 65.4 & 63.3 \\
    {\scriptsize MinMaxCAM}~\cite{wang2021minmaxcam} & - & 66.7 & 65.7 \\
    \hline
    \modelname$_\text{Shared}$ & \textbf{60.2} & \textbf{68.1} & \textbf{67.8} \\
    \hline
\end{tabular}
}
\label{table:loc}
\end{minipage}%
\qquad
\begin{minipage}{.48\textwidth}

\centering
\caption{\modelname~pre-training with additional image-text pairs. ``Separate$^{\dagger\dagger}$'' uses additional 4M image-text pairs from CC~\cite{sharma2018conceptual} and SBU~\cite{ordonez2011im2text} that do not have grounded box annotations.} 
\resizebox{0.95\textwidth}{!}
{
\begin{tabular}{ l | c c | c | c }
    \hline
    \multirow{2}{*}{\modelname} & \multicolumn{2}{c|}{Grounded caption} & COCO & VQAv2 \\
      & Cider & F1$_{all}$ & test-Cider & KP-test \\
    \hline
    Separate & 65.6 & 11.46 & 119.3 & 66.6 \\ 
    Separate$^{\dagger\dagger}$ & 74.2 & 12.62 & 123.1 & 69.1 \\ 
    \hline
\end{tabular}
}
\label{table:text}
\end{minipage}
\end{table}

%% file: conclusion.tex
\section{Conclusion}
We have presented \modelname~that unifies text and box outputs for grounded VL modeling. With the special $\tiny{<}obj\tiny{>}$ token, \modelname~could generate both text and box predictions, with the word-box alignments naturally represented in the output sequence. Unifying text and box outputs allows the model to better approach grounded VL tasks such as grounded captioning. Furthermore, the unified multi-task network and the task-agnostic output sequence design make \modelname~parameter efficient and generalizable to new tasks. We see great potential in \modelname, and believe it paves the way for building vision systems with stronger intelligence, such as in-context learning~\cite{brown2020language} and instruction tuning~\cite{wei2021finetuned}.

%% file: supply.tex
\appendix
\section{Experiment Details}
\label{sec:supp_setup}
\noindent\textbf{Hyper-parameter.}
We summarize the detailed experiment settings of \modelname~in Table~\ref{table:param}. In the \modelname~decoder, we encode previous target token inputs $s_{<t}$ with token and position embedding, and do not use type embedding to differentiate text and box tokens. 

\noindent\textbf{Training corpus.}
We introduce the ``200K'' pre-training corpus in the main paper, which contains both image-text pairs and grounded box annotations. In the main paper's Tables 4-6, we refer to the training corpus used in previous studies by their contained image numbers. Specifically, the ``180K'' corpus~\cite{cho2021unifying,xu2021e2e} aggregate images and annotations from COCO~\cite{lin2014microsoft} and Visual Genome~\cite{krishna2017visual}. 

The ``3M'' corpus~\cite{zhou2020unified,lu2019vilbert} contains image-text pairs from the Conceptual Captions dataset~\cite{sharma2018conceptual}.  The ``4M'' corpus~\cite{chen2019uniter,li2020oscar,gan2020large} consists of the COCO~\cite{lin2014microsoft}, Visual Genome~\cite{krishna2017visual}, Conceptual Captions~\cite{sharma2018conceptual}, and SBU Captions~\cite{ordonez2011im2text} image-text pairs.

\noindent\textbf{Downstream task post-processing and evaluation.}
We detail the post-processing and downstream task evaluation in \modelname~inference. The first step shared among different tasks is to extract text, box, and word-box alignment predictions from the unified output sequence, as visualized in the main paper's Figure 3. We then use the extracted outputs for downstream task evaluations.
We next detail the evaluation process of specific downstream tasks. \textbf{1). Grounded captioning.} We use the extracted text, box, and alignment predictions to compute caption and grounding evaluations following the standard benchmark~\cite{zhou2019grounded}. \textbf{2). Phrase grounding.} We require the model to repeat the input text query and ground boxes as box tokens inline in the output sequence. For phrase grounding, the model needs to predict object boxes and align the box with words in the input text query. Instead of separately predicting the alignments between predicted boxes and input words~\cite{kamath2021mdetr,harold_GLIP2021}, \modelname~repeats the input text and extracts alignments with the $\tiny{<}obj\tiny{>}$ token from the unified output sequence. 
If the repeated text output is wrong, the alignment will be disarrayed, thus leading to wrong phrase grounding predictions. \textbf{3). Referring expression comprehension.} Since the referring expression comprehension task~\cite{yu2016modeling,mao2016generation} does not require the alignment prediction, we take the first predicted box in the output sequence as the final grounding prediction. \textbf{4). COCO captioning and VQA.} We use the extracted text outputs for final evaluations (\ie, captioning metrics for COCO and exact match for VQA accuracy).

\input{tabs/supply/param}

\section{Ablation Studies on Decoding Design}
\label{sec:supp_ablation}
In this section, we present ablation studies on \modelname~decoder and output sequence design, starting with ``$\tiny{<}obj\tiny{>}$ token,'' ``decoder type embedding,'' and ``number of object tokens.'' For these three ablation studies on the decoder, we initialize single-modality and transformer encoders with pre-trained \modelname~weights, and finetune model variants that have different decoder designs on the experimented task. We then discuss different inference-time ``decoding sampling method,'' and the experiment on ``decoding syntactic restrictions.''

\noindent\textbf{$\mathbf{\tiny{<}obj\tiny{>}}$ token.}
\modelname's special $\tiny{<}obj\tiny{>}$ token naturally represents the word-box alignments in the output sequence. In addition to indicating the alignments, we examine if $\tiny{<}obj\tiny{>}$ simplifies the sequence prediction and thus improves the model performance. The referring expression comprehension task requires a single box output and does not need the word-box alignment. Thus, we could remove the $\tiny{<}obj\tiny{>}$ token while still being able to perform the task.
Table~\ref{table:objtoken} shows the experiments on the Refcocog dataset~\cite{mao2016generation}. The \modelname$_\text{Separate}$ baseline inserts a pair of $\tiny{<}obj\tiny{>}$ and ${{\tiny<}{\tiny\backslash}\text{obj}{\tiny>}}$ tokens before and after a word-box token segment. We experiment with removing the ${{\tiny<}{\tiny\backslash}\text{obj}{\tiny>}}$ token, or both special tokens. We observe an around $1\%$ accuracy improvement by adding $\tiny{<}obj\tiny{>}$ tokens.

\input{tabs/supply/obj_ablation}

\noindent\textbf{Decoder type embedding.}
Table~\ref{table:typeemd} shows the ablation study on decoder type embedding, \ie, whether to use type embedding to differentiate text and box tokens. We experiment with the following variants of decoder embedding~\cite{hu2020iterative,yang2021tap}. The \modelname$_\text{Separate}$ baseline does not use type embedding. ``${{\tiny<}\text{obj}{\tiny>}}$ as text/box tokens'' uses two type embedding to differentiate text and box tokens, where the ${{\tiny<}\text{obj}{\tiny>}}$ token is tagged as text or box token. ``${{\tiny<}\text{obj}{\tiny>}}$ as a third type'' introduces an extra type embedding specialized for ${{\tiny<}\text{obj}{\tiny>}}$ and ${{\tiny<}{\tiny\backslash}\text{obj}{\tiny>}}$. We experiment on the Refcocog~\cite{mao2016generation} and Flickr30k Entities~\cite{plummer2015flickr30k} grounding tasks. We empirically observe that the decoder type embedding has no major influence on model performance, and thus do not use type embedding in \modelname.

\begin{figure}[t]
\begin{minipage}{.58\textwidth}
  \centering
  \includegraphics[width=\textwidth]{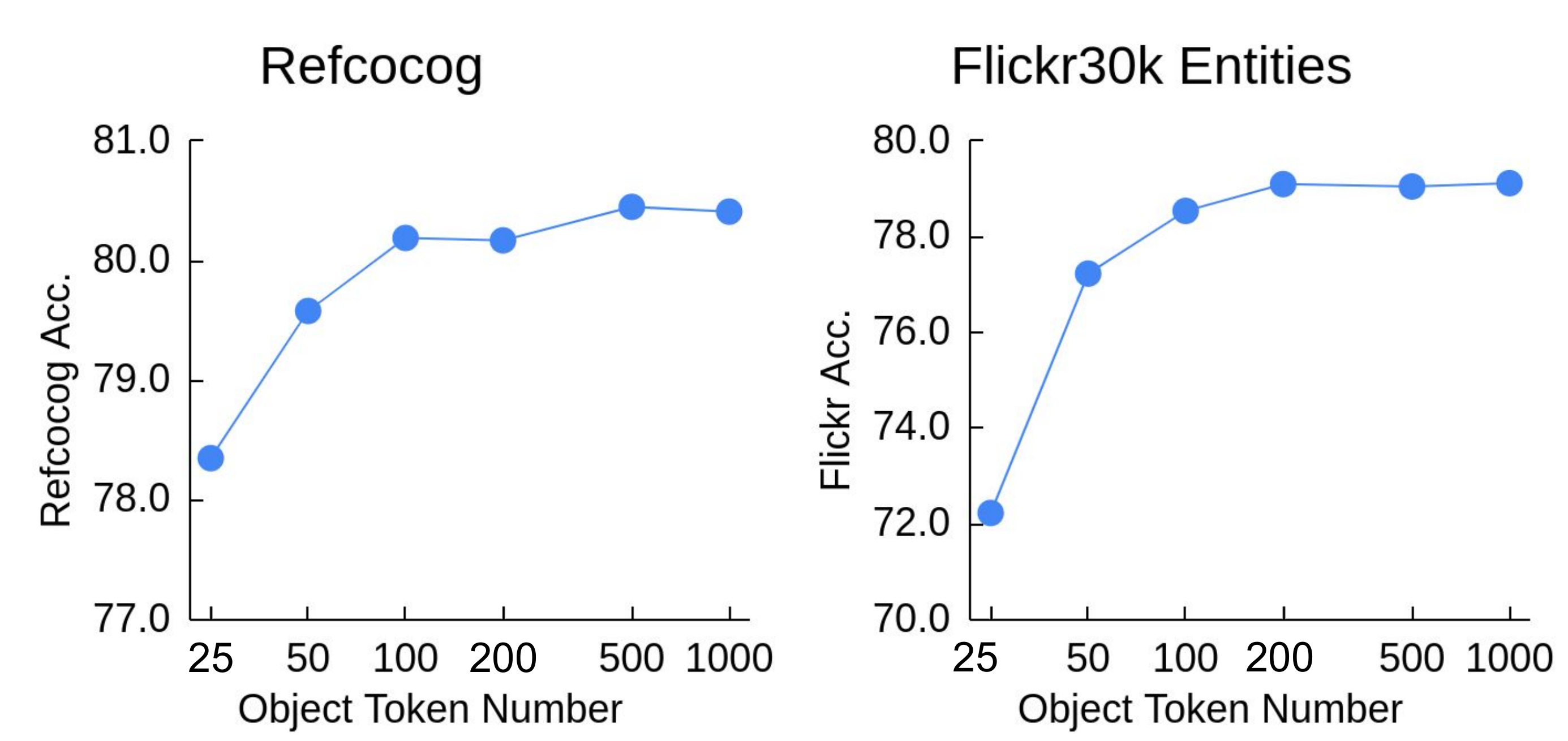}
  \caption{Ablation studies on the box token number. We experiment on the Refcocog~\cite{mao2016generation} and Flickr~\cite{plummer2015flickr30k} grounding tasks.}
  \label{fig:bins}
\end{minipage}%
\qquad
\begin{minipage}{.37\textwidth}
  \centering
  \includegraphics[width=\textwidth]{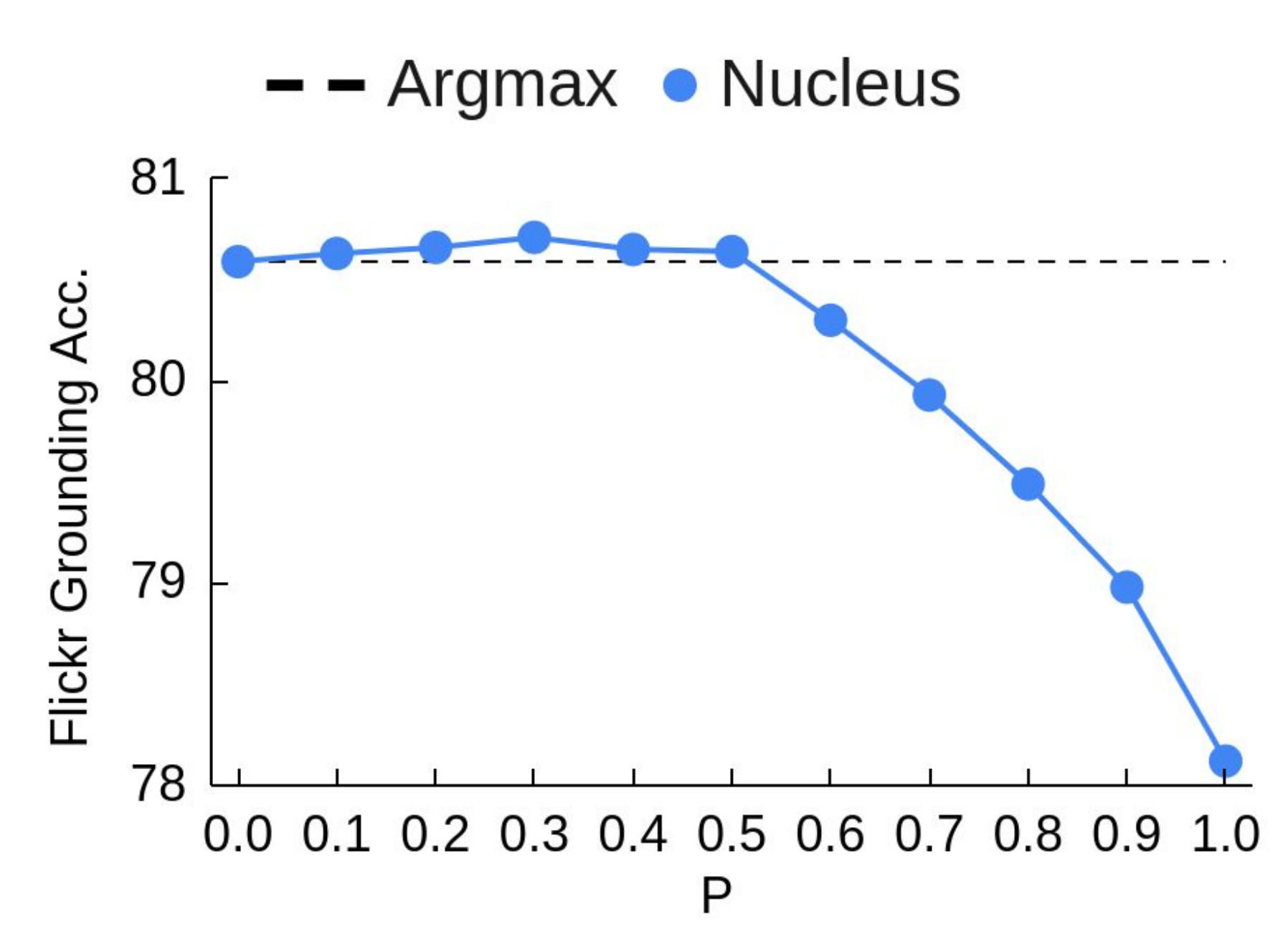}
  \caption{Ablation studies on decoding sampling method on the phrase grounding task.}
  \label{fig:decode}
\end{minipage}
\end{figure}

\noindent\textbf{Number of object tokens.}
Figure~\ref{fig:bins} shows the influence of object token number on the grounding performance. We observe a steady performance when the object token number is large enough for a dataset to avoid quantization error. The token number is around 200 for the experimented VL datasets.

\noindent\textbf{Decoding sampling method.}
Figure~\ref{fig:decode} shows the ablation study of the decoding sampling method on Flickr phrase grounding~\cite{plummer2015flickr30k}. Compared with the simple argmax decoding sampling, we observe a marginal improvement from nucleus sampling~\cite{holtzman2019curious,chen2021pix2seq}. The improvement from nucleus sampling is smaller on \modelname's experimented VL tasks, compared with previous explorations on object detection~\cite{chen2021pix2seq}. We suspect that the smaller gain is due to the difference in target sequences. Specifically, object detection~\cite{chen2021pix2seq} has multiple correct decoding sequences, as object order doesn't matter in object detection outputs. In contrast, \modelname~only has one fixed decoding target of the constructed text+box sequence. Thus, the diversity brought by nucleus sampling helps less in \modelname.

\noindent\textbf{Decoding syntactic restriction.}
We apply no decoding syntactic restriction in \modelname~ training.
We scan \modelname~predictions for two types of failure cases that break the syntactic restrictions in output decoding sequences: \textbf{(1)} before $\tiny{<}\backslash obj\tiny{>}$ there are not exactly four consecutive box tokens; \textbf{(2)} $\tiny{<} obj\tiny{>}$, $\tiny{<}\backslash obj\tiny{>}$ tokens are followed by box tokens, or are not paired. We scan the Refcocog grounding and Flickr grounded captioning predictions generated by \modelname$_\text{Pre-finetuning}$, and COCO captions generated by \modelname$_\text{Shared}$ (for generalized grounded captioning in Figure~3~(d)). We observe \textbf{zero} syntactic failure cases in all scanned outputs, implying that the decoding token type pattern is easy to learn.

We then incorporate these two syntactic restrictions into the model training, and examine if the restrictions ease the training and improve model performance. Specifically, we compute the softmax language modeling loss over a subset of all tokens in applicable decoding positions, such as masking out box logits after $\tiny{<} obj\tiny{>}$. We experiment on Refcoco and Flickr grounded captioning, based on \modelname$_\text{Separate}$. We empirically observe that the syntactic restriction has no major influence on the model performance, with $+0.9$ accuracy gain on Refcoco and a slight drop on Flickr grounded captioning ($-0.2$ CIDEr, $-0.15$ $F1_{all}$).

\section{Discussions}
\noindent\textbf{Pre-training with additional box annotations.}
\input{tabs/supply/oddata}
In addition to the ``200K'' pre-training corpus and the additional image-text data~\cite{sharma2018conceptual,ordonez2011im2text} introduced in the main paper, we further explore using additional box annotations with no caption texts for \modelname~pre-training. We aggregate object detection annotations from  COCO~\cite{lin2014microsoft}, VG~\cite{krishna2017visual}, Objects365~\cite{shao2019objects365}, and OpenImages~\cite{kuznetsova2020open}. Each sample is an image with object box and class annotations. For pre-training, we randomly select up to $32$ objects and shuffle the object order. We concatenate the object class name as the input text, and train the model to generate the text+box sequence to ground the selected objects. We refer to \modelname$_\text{Separate}$ with those extra box annotations as ``Separate$^\text{box}$.''

Table~\ref{table:oddata} shows the experiment results of adding additional box annotations. On VL tasks that require box prediction, such as the visual grounding task and the grounding evaluation in grounded captioning, ``Separate$^\text{box}$'' consistently outperforms \modelname$_\text{Separate}$ on the grounding accuracy and grounded captioning F1 score. More interestingly, we empirically observe that extra box annotations could also help the text output quality. For example, ``Separate$^\text{box}$'' improves grounded captioning CIDEr score from $65.6$ to $70.0$, COCO captioning CIDEr score from $119.3$ to $120.7$, and VQA accuracy from $66.6\%$ to $68.4\%$.

\noindent\textbf{Multi-task finetuning with prefix.}
\input{tabs/supply/prefix}
In the main paper, we discuss the effectiveness of multi-task finetuning, which gathers training data from all downstream tasks and learns a single set of parameters for different VL tasks. By unifying all considered downstream tasks as a sequence generation problem, a single \modelname$_\text{Shared}$ model could perform well on different tasks, meanwhile being parameter efficient and showing promise in zero-shot generalization.

One variant of \modelname$_\text{Shared}$ is to add a task-specific input text string to identify the task for each sample~\cite{cho2021unifying}, such as ``visual grounding:''. The extra input text string is known as the prefix. We experiment with a variant of \modelname~multi-task finetuning with prefix, namely \modelname$_\text{Prefix}$. \modelname$_\text{Prefix}$ adds a task-specific prefix at the beginning of each input text string, \eg, ``Visual grounding: the coffee mug next to the plate.'' We use the task name as the prefix, \ie, ``visual grounding:'', ``phrase grounding:'', ``grounded captioning:'', ``image captioning:'', ``question answering:'', \etc. We then train the model with multi-task finetuning, the same as \modelname$_\text{Shared}$.
Table~\ref{table:prefix} compares \modelname$_\text{Prefix}$ with \modelname$_\text{Shared}$. We observe a comparable performance with and without prefix on the experimented tasks and datasets.

\input{tabs/supply/bias}
\noindent\textbf{Robustness and bias analyses.}
We conduct robustness and bias analyses to better understand the limitation of \modelname. Tables~\ref{table:ana_vqa},\ref{table:ana_cap} show initial robustness and bias analyses. We retrain \modelname~on the splits (VQA-train, COCO-14) used in the established analysis setups~\cite{hendricks2018women,dancette2021beyond}. In Table~\ref{table:ana_vqa}, we follow VQA-CE~\cite{dancette2021beyond} and compare the gain over the UpDown baseline on two subsets. \modelname~achieves a larger gain on ``counterexamples'' ($+9.76\%$) compared with ``easy'' ($+6.17\%$), indicating better robustness against shortcuts compared with the UpDown baseline, as discussed in VQA-CE~\cite{dancette2021beyond}. Table~\ref{table:ana_cap} evaluates gender bias with error rate~\cite{hendricks2018women}. \modelname~achieves a lower error rate than general captioning models (\cf, Baseline-FT of $19.30\%$ \vs \modelname~of $9.21\%$), and is only slightly worse than the specialized method Equalizer~\cite{hendricks2018women}. We hypothesize that the reasonable robustness and bias performance is related to ~\modelname's grounded training, which better binds visual concepts with text words.
Despite the reasonable performance on the standard analyses, building robust and unbiased models remains a challenging problem and could be further improved. 

\section{Qualitative Results}
\label{sec:sup_visu}

\begin{figure*}[t]
\centering
\includegraphics[width=1.\textwidth]{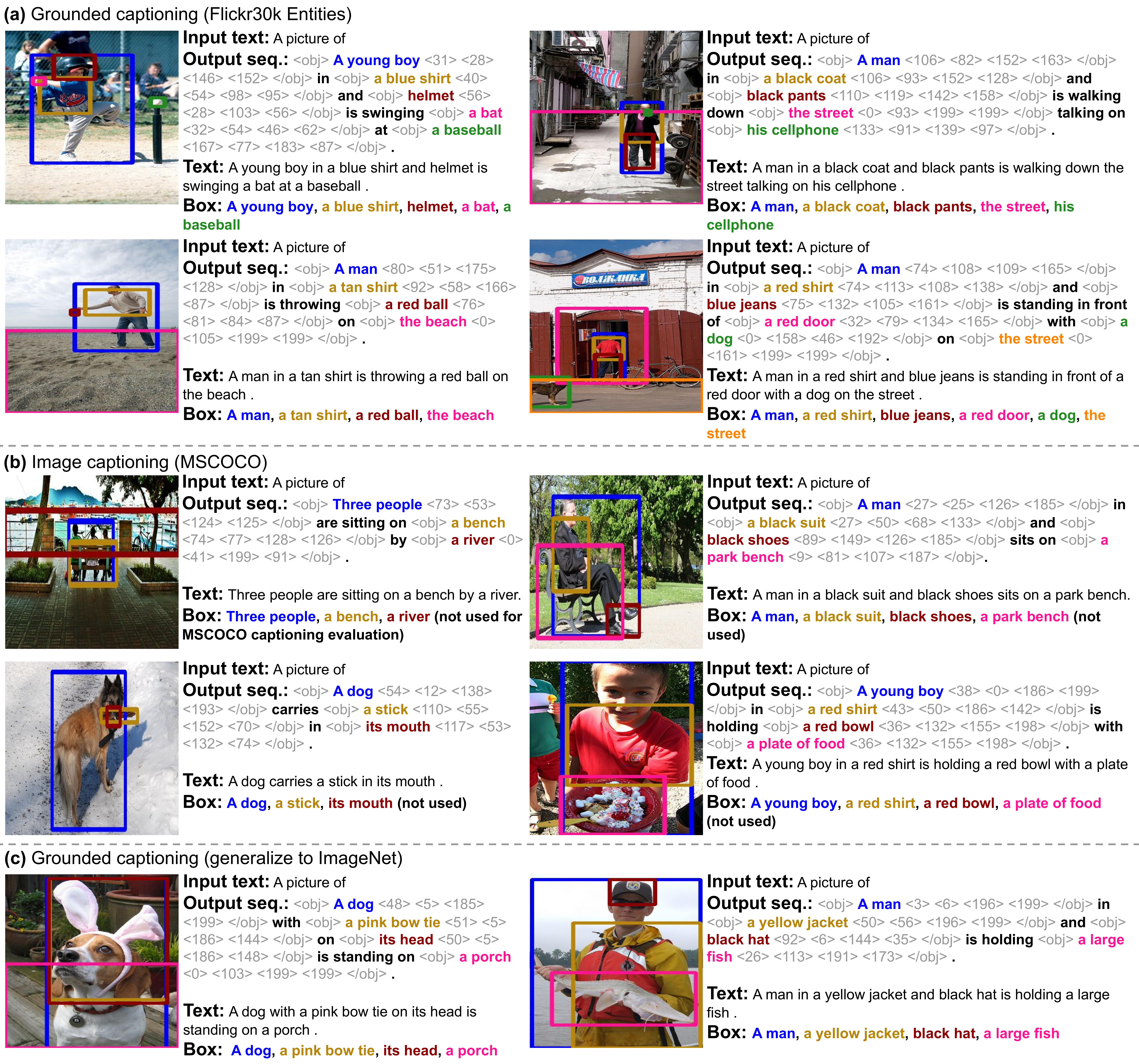}
    \caption{
    Additional qualitative results from \modelname$_\text{Shared}$ on captioning tasks.
	}
\label{fig:suppB}
\end{figure*}
\begin{figure*}[t]
\centering
\includegraphics[width=1.\textwidth]{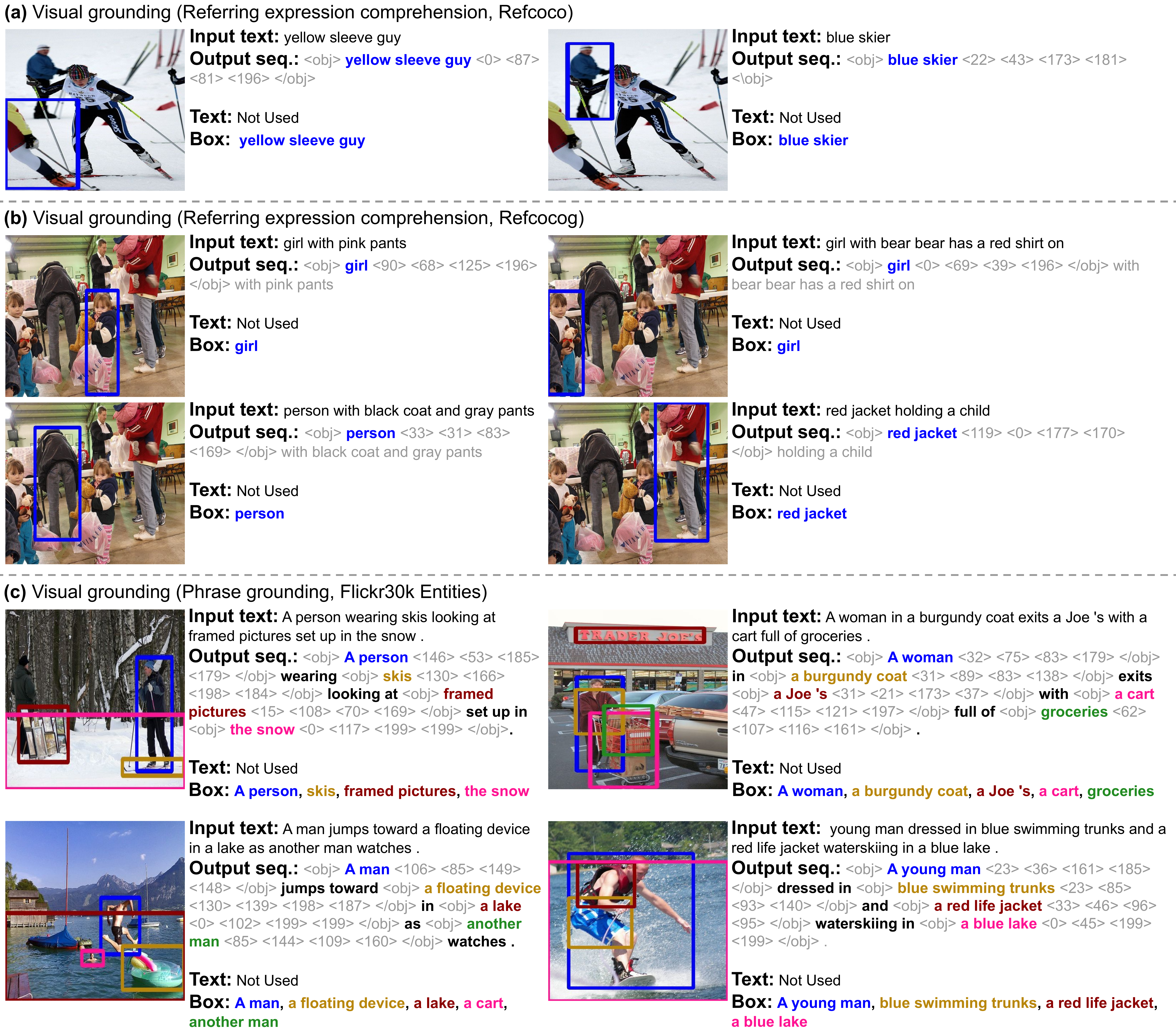}
    \caption{
    Additional qualitative results from \modelname$_\text{Shared}$ on grounding tasks.
	}
\label{fig:suppA}
\end{figure*}

In this section, we present additional qualitative results made by \modelname$_\text{Shared}$. We start with the captioning tasks in Figure~\ref{fig:suppB}. Figure~\ref{fig:suppB}\textbf{(a)} presents the grounded captioning results on Flickr30k Entities, where the predictions are evaluated by both the caption and grounding metrics. \modelname~performs well in both generating captions and grounding noun phrases to image regions. For captioning, the model generates a smooth and accurate image description, and properly includes attribute words to produce an informative caption, \eg, ``young boy'' and ``blue shirt'' in the top left example. \modelname~is also capable of providing a comprehensive description of the scene. For example, in the bottom right sub-figure of (a), the caption consists of the foreground object and its detailed attributes ``man in red shirt and blue jeans,'' scene descriptions ``a red door'' and ``on the street,'' and the nearby object ``a dog.'' The model also performs well in grounding. Noticeably, \modelname~performs well on grounding and describing tiny objects, \eg, the ``a bat'' and ``a baseball'' in the top left example and the ``a red ball'' in the bottom left example.

Figure~\ref{fig:suppB}\textbf{(b)} shows \modelname's prediction on the MSCOCO image captioning task. With the same inputs as Flickr30k grounded captioning, \modelname~learns to transfer the grounded captioning ability learned on Flirckr30k Entities to MSCOCO, although COCO captioning does not have grounding annotations. For evaluation, we extract the text tokens and compute the standard COCO captioning metrics~\cite{papineni2002bleu,denkowski2014meteor,anderson2016spice,vedantam2015cider}. We note that \modelname~achieves comparable caption performance to the state of the art, and meanwhile being capable of grounding all noun phrases in the caption.
As shown in Figure~\ref{fig:suppB}(b), \modelname~generates an informative caption and accurately grounds all noun phrases in the caption to visual regions. Such grounded captioning ability is important for reducing object hallucination~\cite{rohrbach2018object}, boosting the model's interpretability and fairness~\cite{rohrbach2018object,hendricks2018women}, and facilitating applications in robotics and human-computer interaction.
We also visualize additional captioning examples on ImageNet~\cite{deng2009imagenet}. We observe that \modelname~generalizes well onto the ImageNet images. The ImageNet caption and grounding predictions in Figure~\ref{fig:suppB}\textbf{(c)} are of similar qualities as on Flickr30k Entities and MSCOCO.

Figure~\ref{fig:suppA} shows \modelname$_\text{Shared}$'s predictions on grounding tasks. Figures~\ref{fig:suppA}(a,b) are from the Refcoco~\cite{yu2016modeling} and Refcocog~\cite{mao2016generation} datasets, for the referring expression comprehension task. We observe that the model learns to identify different objects in the same image conditioned on different input queries. For example, in Figure~\ref{fig:suppA}\textbf{(a)}, the targets of ``yellow sleeve guy'' on the left and ``blue'' skier in the background. Similarly, in Figure~\ref{fig:suppA}\textbf{(b)}, \modelname~correctly differentiates the four people in the image. \modelname~also correctly localizes the head noun in a long referring expression and predicts the box on the corresponding phrase. For example, in Figure~\ref{fig:suppA}(b), grounding boxes are predicted on the words ``girl'' and ``person,'' instead of the entire query as in previous studies~\cite{yu2018mattnet,yang2019fast,deng2021transvg}. 
Another observation is that \modelname~usually predicts a single box in the output sequence for referring expression comprehension samples. For example, in the top left sub-figure of Figure~\ref{fig:suppA}(b), the model only grounds the head noun ``girl'' and does not generate a box for the remaining phrase like ``pink pants.'' We conjecture that \modelname~learns to identify the referring expression comprehension task based on the input text (\eg, a short referring query \vs a complete sentence), and generates a single box when performing the task. 

Figure~\ref{fig:suppA}\textbf{(c)} shows the phrase grounding examples on the Flickr30k Entities dataset~\cite{plummer2015flickr30k}. Phrase grounding requires the model to identify all noun phrases in a sentence and ground them to corresponding image regions. \modelname~correctly grounds all types of phrases referred to in the sentence, including foreground objects ``person'' and ``woman,'' smaller background objects ``skies'' in the top left example and ``another man'' in the bottom left example, and scene regions ``the snow,'' ``a lake,'' and ``a blue lake.'' The model even correctly predicts challenging regions such as the ``trader joe's'' logo in the top right sub-figure.


%% file: tabs/supply/param.tex
\begin{table}[t]
\centering
\caption{The detailed experiment settings of \modelname.}
\resizebox{0.61\textwidth}{!}
{
\begin{tabular}{ l c}
    \hline
    Hyper-parameter & Value \\
    \hline
    \textcolor{blue}{(a) Optimizer hyper-parameters} & \\
    optimizer & AdamW~\cite{loshchilov2017decoupled} \\
    base learning rate & 1e-4 \\
    backbone learning rate & 2e-5 \\
    learning rate schedule & Step *0.1 for final 5 epochs \\
    weight decay & 1e-4 \\
    batch size & 64 \\
    pre-training epochs & 40 \\
    multi-task finetuning epochs & 20 \\
    task-specific finetuning epochs & 20 \\
    exp. moving average & 0.9998 \\
    \hline
    \textcolor{blue}{(b) Model hyper-parameters} & \\
    encoder layer number & 6 \\
    encoder hidden size & 256 \\
    encoder intermediate size & 2048 \\
    encoder head number & 8 \\
    decoder layer number & 6 \\
    decoder hidden size & 256 \\
    decoder intermediate size & 2048 \\
    decoder head number & 8 \\
    max input words & 256 \\
    input visual tokens & $\frac{H_0}{32}\times\frac{W_0}{32}$~\cite{he2016deep} \\
    max decoding steps & 256 \\
    number of bins & 200 \\
    augmentation & RandomResizedCrop~\cite{carion2020end} \\
    image size & 800-1333 \\
    encoder vocab size & 50265 (RoBERTa~\cite{liu2019roberta}) \\
    decoder vocab size & 50265+200+2=50467 \\
    \hline
\end{tabular}
}
\label{table:param}
\end{table}

%% file: tabs/supply/obj_ablation.tex
\begin{table}[t]\small
\begin{minipage}{.38\textwidth}
\centering
\caption{Ablation study of the $\mathbf{\tiny{<}obj\tiny{>}}$ token on the Refcocog~\cite{mao2016generation} dataset.}
\resizebox{1.\textwidth}{!}
{
\begin{tabular}{ l | c}
    \hline
    \modelname$_\text{Separate}$ & Refcocog \\
    \hline
    Baseline & 80.23 \\
    w/o ${{\tiny<}{\tiny\backslash}\text{obj}{\tiny>}}$ & 79.41 \\
    w/o ${{\tiny<}\text{obj}{\tiny>}}$, ${{\tiny<}{\tiny\backslash}\text{obj}{\tiny>}}$ & 79.31 \\
    \hline
\end{tabular}
}
\label{table:objtoken}
\end{minipage}%
\qquad
\begin{minipage}{.55\textwidth}

\centering
\caption{Ablation study of the decoder type embedding. We experiment on the Refcocog~\cite{mao2016generation} and Flickr30k Entities~\cite{plummer2015flickr30k} grounding tasks.}
\resizebox{1.\textwidth}{!}
{
\begin{tabular}{ l | c c}
    \hline
    \modelname$_\text{Separate}$ & Refcocog & Flickr \\
    \hline
    Baseline (w/o type emd.) & 80.23 & 78.92 \\
    ${{\tiny<}\text{obj}{\tiny>}}$ as text tokens & 80.47 & 78.65 \\
    ${{\tiny<}\text{obj}{\tiny>}}$ as box tokens & 80.39 & 79.40 \\
    ${{\tiny<}\text{obj}{\tiny>}}$ as a third type & 80.54 & 78.83 \\
    \hline
\end{tabular}
}
\label{table:typeemd}
\end{minipage}
\end{table}

%% file: tabs/supply/oddata.tex
\begin{table*}[t]\small
\centering
\caption{\modelname~pre-training with additional bounding box annotations. ``Separate$^\text{box}$''
adopts the extra bounding box annotations from the COCO~\cite{lin2014microsoft}, VG~\cite{krishna2017visual}, Objects365~\cite{shao2019objects365}, OpenImages~\cite{kuznetsova2020open} object detection datasets.
}
\resizebox{1.\textwidth}{!}{
\begin{tabular}{ l || c c c c | c c | c | c}
    \hline
    \multirow{2}{*}{\modelname} &
    \multicolumn{4}{c|}{Visual grounding} & \multicolumn{2}{c|}{Grounded caption} & COCO & VQAv2\\
     & Refcoco & Refcoco+ & Refcocog & Flickr & Cider & F1$_{all}$ & test-Cider & KP-test \\
    \hline
    Separate & 86.32 & 78.70 & 79.96 & 79.39 & 65.6 & 11.46 & 119.3 & 66.6 \\
    Separate$^\text{box}$ & 88.27 & 80.98 & 83.78 & 81.90 & 70.0 & 13.46 & 120.7 & 68.4 \\
    \hline
\end{tabular}
}
\label{table:oddata}
\end{table*}

%% file: tabs/supply/prefix.tex
\begin{table*}[t]\small
\centering
\caption{Experiment results of \modelname$_\text{Prefix}$ that adds task-specific prefix in multi-task finetuning.
}
\resizebox{1.\textwidth}{!}{
\begin{tabular}{ l | c || c c c c | c c | c | c c}
    \hline
    \multirow{2}{*}{Method} & \#Pre- &
    \multicolumn{4}{c|}{Visual grounding} & \multicolumn{2}{c|}{Grounded caption} & COCO & {VQAv2}\\
     & train & Refcoco & Refcoco+ & Refcocog & Flickr & Cider & F1$_{all}$ & test-Cider & KP-test \\
    \hline
    MDETR~\cite{kamath2021mdetr} & 200K & 86.75 & {79.52} & 81.64 & 83.8 & - & - & - & - \\
    UNITER~\cite{chen2019uniter} & 4M & 81.24 & 75.31 & 74.31 & - & - & - & - & {70.5} \\
    GVD~\cite{zhou2019grounded} & - & - & - & - & - & 62.3 & 7.55 & - &  - \\
    VL-T5~\cite{cho2021unifying} & 180K & - & - & 71.2 & - & - & - & 116.5 & 67.9 \\
    OSCAR~\cite{li2020oscar} & 4M & - & - & - & - & - & - & {123.7} & - \\
    \hline
    \modelname$_\text{Shared-Scratch}$ & None & 82.06 & 70.72 & 73.39 & 65.67 & 61.1 & 7.85 & 111.8 & 63.1 \\
    \modelname$_\text{Prefix-Scratch}$ & None & 82.38 & 70.96 & 75.43 & 69.58 & 62.1 & 8.51 & 112.8 & 64.3 \\
    \rowcolor{LightGray}
    \modelname$_\text{Shared}$ & 200K & 88.50 & 80.98 & 84.46 & 79.23 & 63.4 & 9.18 & 115.8 & 66.6 \\
    \rowcolor{LightGray}
    \modelname$_\text{Prefix}$ & 200K & 87.60 & 79.72 & 83.41 & 80.13 & 62.4 & 10.54 & 115.6 & 66.0 \\
    \hline
\end{tabular}
}
\label{table:prefix}
\end{table*}

%% file: tabs/supply/bias.tex
\begin{table}[t]\small
\begin{minipage}{.5\textwidth}
\centering
\caption{VQA-CE~\cite{dancette2021beyond} robustness analyses.}
\resizebox{1.\textwidth}{!}
{
\begin{tabular}{ l c c }
    \hline
    Accuracy(\%) & UpDown & \modelname \\
    \hline
    Overall & 63.52\scriptsize{~\textcolor{darkgreen}{(+0.00)}} & 70.78\scriptsize{~\textcolor{darkgreen}{(+7.26)}} \\
    Counterexamples & 33.91\scriptsize{~\textcolor{darkgreen}{(+0.00)}} & 43.67\scriptsize{~\textcolor{darkgreen}{(+9.76)}} \\
    Easy & 76.69\scriptsize{~\textcolor{darkgreen}{(+0.00)}} & 82.86\scriptsize{~\textcolor{darkgreen}{(+6.17)}} \\
    \hline
\end{tabular}
}
\label{table:ana_vqa}
\end{minipage}%
\qquad
\begin{minipage}{.45\textwidth}
\centering
\caption{Gender error analyses~\cite{hendricks2018women}.}
\resizebox{1.\textwidth}{!}
{
\begin{tabular}{ l c c }
    \hline
    Error rate(\%) & \scriptsize{COCO-Bias} & \scriptsize{COCO-Balanced} \\
    \hline
    BaselineFT & 12.83 & 19.30 \\
    Balanced & 12.85 & 18.30 \\
    UpWeight & 13.56 & 16.30 \\
    Equalizer & 7.02 & 8.10 \\
    \modelname & 9.87 & 9.21 \\
    \hline
\end{tabular}
}
\label{table:ana_cap}
\end{minipage}
\end{table}